\documentclass[lettersize,journal]{IEEEtran}
\usepackage{amsmath,amsfonts}
\usepackage{algorithmic}
\usepackage{algorithm}
\usepackage{array}
\usepackage[caption=false,font=normalsize,labelfont=sf,textfont=sf]{subfig}
\usepackage{textcomp}
\usepackage{stfloats}
\usepackage{url}
\usepackage{verbatim}
\usepackage{graphicx}
\usepackage{wrapfig} 
\usepackage{lipsum}
\usepackage{graphicx}
\usepackage{booktabs} 
\usepackage{threeparttable} 
\graphicspath{{./images/}}
\usepackage{array}       
\usepackage[
  style=numeric,        
  sorting=none,        
  backend=biber,       
  giveninits=true,     
  maxnames=6           
]{biblatex}
\DeclareCaptionFont{subfont}{\normalfont} 
\begin{document}

\title{Fusing Driver Perceived and Physical Risk for Safety Critical Scenario Screening in Autonomous Driving}

\author{Chen Xiong, Ziwen Wang, Deqi Wang, Cheng Wang, Yiyang Chen, He Zhang\textsuperscript{†}, Chao Gou 
\thanks{† Corresponding author (zhanghe@swjtu.edu.cn.)}
\thanks{Chen Xiong, Ziwen Wang, Deqi Wang, Cheng Wang, Yiyang Chen and Chao Gou are with the Guangdong Provincial Key Laboratory of Intelligent Transportation System, School of Intelligent Systems Engineering, Shenzhen Campus of Sun Yat-sen University, Shenzhen, China}
\thanks{He Zhang is with the school of Transportation and Logistics, Southwest Jiaotong University, Chengdu, China.}
\thanks{The authors would like to thank the fundings of the National Key R\&D Program of China (Grant No. 2024YFB4303400), Natural Science Foundation of Guangdong Province, China (Grant No. 2025A1515010166), Shenzhen Fundamental Research Program, China (Grant No. JCYJ20240813151301003) and Autonomous Transportation Collaborative Innovation Platform of Longhua District, Shenzhen, China.}}

\markboth{IEEE Intelligent Transportation Systems Magazine}%
{Shell \MakeLowercase{\textit{et al.}}: A Sample Article Using IEEEtran.cls for IEEE Journals}

\IEEEpubid{}

\maketitle

\begin{abstract}
Autonomous driving testing increasingly relies on mining safety critical scenarios from large scale naturalistic driving data, yet existing screening pipelines still depend on manual risk annotation and expensive frame by frame risk evaluation, resulting in low efficiency and weakly grounded risk quantification. To address this issue, we propose a driver risk fusion based hazardous scenario screening method for autonomous driving. During training, the method combines an improved Driver Risk Field with a dynamic cost model to generate high quality risk supervision signals, while during inference it directly predicts scenario level risk scores through fast forward passes, avoiding per frame risk computation and enabling efficient large scale ranking and retrieval. The improved Driver Risk Field introduces a new risk height function and a speed adaptive look ahead mechanism, and the dynamic cost model integrates kinetic energy, oriented bounding box constraints, and Gaussian kernel diffusion smoothing for more accurate interaction modeling. We further design a risk trajectory cross attention decoder to jointly decode risk and trajectories. Experiments on the INTERACTION and FLUID datasets show that the proposed method produces smoother and more discriminative risk estimates. On FLUID, it achieves an AUC of 0.792 and an AP of 0.825, outperforming PODAR by 9.1 percent and 5.1 percent, respectively, demonstrating its effectiveness for scalable risk labeling and hazardous scenario screening.
\end{abstract}


\section{Introduction}
\IEEEPARstart{I}{ntelligent} connected and autonomous driving technologies are advancing rapidly, with safety and reliability becoming critical prerequisites for industrial deployment \cite{sun2021scenario},\cite{bian2025search}. Effective validation of autonomous driving systems relies on extensive real-world driving scenarios, particularly rare yet extremely hazardous edge cases \cite{watanabe2019scenario},\cite{song2022scenario}. Real-world operational data indicates that while severe accidents often concentrate in such scenarios, they constitute an extremely low proportion of natural driving data, exhibiting a typical long-tail distribution \cite{tang2024scenario}. In their research \cite{xu2025wod}, the Waymo team discovered that 11 rare scenarios—such as flocks of birds crossing roads and evading out-of-control motorcycles—selected from 6.4 million miles of real-world driving logs ultimately accounted for only 0.03\% of the total driving data. This scarcity makes it difficult to adequately train and validate autonomous driving systems under critical failure boundaries, constituting a core challenge in current safety testing. Consequently, efficiently and accurately filtering potential hazardous scenarios from vast amounts of natural driving data has become a key pathway for autonomous driving testing and validation.

Currently, scene screening methods can be primarily categorized into two types: the first type relies on Surrogate Safety Measures (SSMs), such as Time to Collision (TTC), Time Headway (THW), and Post-Encroachment Time (PET) \cite{singh2024conflict}. These metrics are computationally straightforward and widely adopted in the industry. However, SSMs typically rely on simplified kinematic assumptions for two vehicles and fixed decision thresholds. They struggle to capture the continuous spatiotemporal evolution of risk and fail to reflect multi-agent interactions and spatiotemporal dynamics in complex traffic environments. Consequently, numerous high-risk samples are obscured within routine data. The second category comprises risk potential field-based methods \cite{kolekar2021risk}, such as driver risk fields, safety fields, and potential collision loss models. These approaches construct two-dimensional potential fields that evolve with vehicle motion, enabling precise physical quantification of risk spatial distribution. However, building high-precision risk fields requires complex frame-by-frame physical iteration calculations. Directly applying them as post processing tools for full-scan analysis of massive open loop data incurs enormous computational costs, failing to meet efficiency demands for scenario screening.

Additionally, when faced with thousands of scenario fragments within vast amounts of natural driving data, conducting exhaustive testing for each scenario would be extremely time-consuming and impractical. Therefore, a risk pre-screener is needed to significantly reduce the workload of identifying high value scenarios. However, building an efficient risk pre-screener requires meeting both accuracy and high-efficiency requirements. The former determines whether it can reliably distinguish potentially high-risk samples, while the latter dictates whether the method can be implemented on large scale datasets.

To overcome these challenges, deep learning-based approaches have gained significant attention. In the field of vehicle trajectory prediction, architectures such as recurrent neural networks (RNNs), long short-term memory (LSTM) networks, and Transformers have been widely adopted due to their ability to capture temporal dependencies and extract spatiotemporal interaction features \cite{katariya2022deeptrack}, \cite{hui2022deep}, \cite{xie2024advdiffuser}. However, directly applying existing trajectory prediction models for hazard screening has limitations, as traditional prediction models 
\begin{figure*}[!t]
\centering
\includegraphics[width=1.0\textwidth]{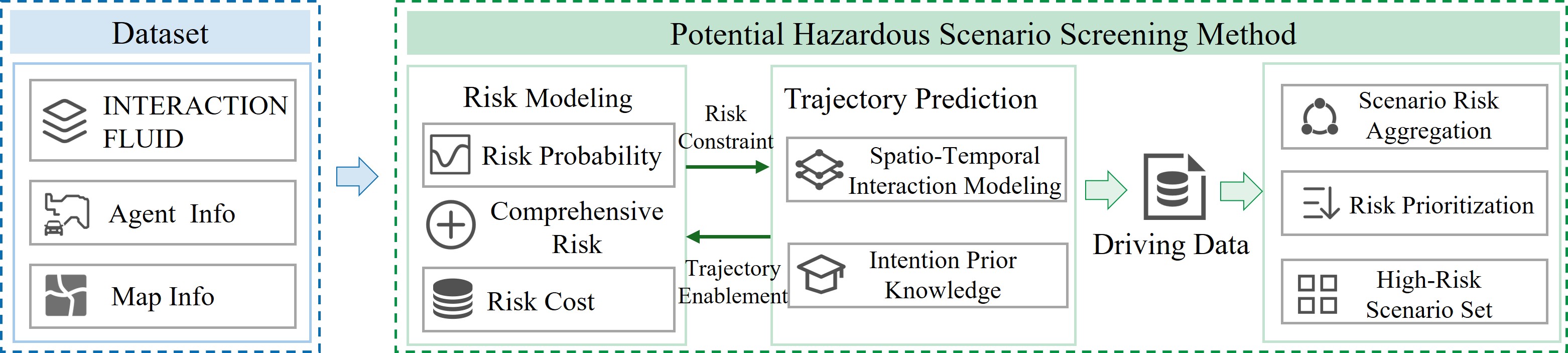}
\caption{Overview of the Problem Formulation. Given an open loop traffic segment with $T$ historical steps and $F$ future steps for $N$ agents, the goal is to infer future risk sequences and aggregate them into a scenario level risk score for prioritizing potentially hazardous segments in large scale naturalistic driving data.}
\label{fig1}
\end{figure*}
typically optimize by minimizing the mean displacement error. This loss function driven approach tends to produce models that closely approximate high probability normal driving behaviors. Consequently, under the long tail distribution of hazardous scenarios, predictions often become overly smooth, which can obscure genuine collision risks. Therefore, purely data driven trajectory prediction cannot replace risk screening mechanisms that incorporate physically grounded prior knowledge.

To address the aforementioned challenges and simultaneously meet the demands for both accuracy and efficiency, we draws inspiration from the intrinsic connection between risk assessment and multi-agent trajectory prediction. On one hand, risk quantification relies on post-processing calculations based on known actual trajectories, resulting in a high degree of coupling between risk supervision signals and trajectory supervision signals in terms of data sources. On the other hand, both require an understanding of the spatiotemporal interaction intentions of multi-agent, exhibiting consistency at the feature representation level. Based on this, we proposes constructing a driving risk based framework for screening potentially hazardous scenarios. By embedding risk assessment capabilities within the prediction model, this framework achieves high speed autonomous screening from data input to hazardous scenarios. Under this framework, trajectory prediction performance is maintained within a reasonable range, enabling it to provide effective kinematic and interaction intent priors for risk prediction.

First, we construct an integrated risk model that combines an enhanced driver risk field with a dynamic cost model. This model is used exclusively during training to compute ground truth risk values for real trajectories, providing the model with high quality, interpretable supervised signals. Second, trajectory prediction is introduced as a secondary task, compelling the model to learn complex spatiotemporal dynamics. This provides crucial prior knowledge about interaction intent for risk prediction. Through multitask joint optimization, the complementary relationship between risk and trajectory is reinforced, enabling the model to remain sensitive to potential risk evolution without explicitly computing the entire future risk profile. Finally, during the screening phase, the model functions as a high speed risk filter. When processing large scale data, it requires only fixed length historical observation segments as input. Its risk branch rapidly infers scenario level risk scores for future time periods based on these segments, while trajectory prediction results serve for interpretability and consistency checks. Thus, the model significantly enhances screening efficiency while achieving efficient filtering from data input to hazardous scenarios.

In summary, the main contributions of this paper are as follows:

1) By refining and integrating the driver risk field with a dynamic cost model, we construct a more stable, comprehensive risk function that aligns better with driver cognition and physical laws. This function captures risk features with greater granularity and foresight than traditional metrics, providing high quality supervisory signals for scenario selection tasks.

2) We propose a risk informed method for screening potentially hazardous scenarios. By embedding risk assessment capability into the prediction model via an implicit input, explicit output risk modeling mechanism and a trajectory risk cross attention module, the proposed approach enables an autonomous pipeline from data ingestion to efficient identification and ranking of hazardous scenarios.

The remainder of this paper is organized as follows: Section 2 reviews prior work on risk assessment and trajectory prediction. Section 3 introduces the proposed methodology, including the construction of an integrated risk model and a network architecture for collaborative decoding of risk trajectories. Section 4 details the experimental setup and evaluation metrics, presenting comparative analyses against multiple baseline methods alongside ablation studies. Section 5 concludes the paper and discusses future research directions.

\section{Related Work}
\subsection{Risk Assessment}
For large scale testing and high value long tail scenario mining in autonomous driving, risk assessment mechanisms must strike an optimal balance between physical interpretability, interaction sensitivity, and data processing scalability. Existing research primarily follows two technical paradigms, kinematic metrics based on proxy safety indicators and continuous risk modeling based on mechanistic potential fields.

Surrogate Safety Measures primarily rely on predefined kinematic and dynamic assumptions to construct collision metrics by calculating spatiotemporal safety intervals. Typical examples include Time to Collision (TTC), Time Headway (THW), Time to Steer (TTS), and Post-Entry Time (PET) \cite{singh2024conflict}, \cite{tafidis2023application}, \cite{wang2022acceleration}, \cite{chang2018integration}. These metrics offer engineering advantages such as formal intuitiveness and low computational overhead, making them highly suitable for low cost preliminary screening of massive datasets. They also demonstrate robust performance in structured environments like highways \cite{cheng2025emergency}. However, SSMs inherently rely on simplified assumptions of pairwise interactions and threshold rules. They often lack sufficient generalization capability when addressing complex urban road scenarios involving multi-agent strong coupling interactions, nonlinear regulatory maneuvers, and occlusions. Their discrete measurement approach struggles to continuously capture the spatiotemporal evolution of risks, making them prone to missing long tail hazardous segments.

Mechanism based continuous risk modeling and potential field modeling aim to provide a fine grained characterization of the spatial distribution of risk and its evolutionary trends from a physical perspective. Such methods often draw upon artificial potential field or energy field theories, mapping surrounding vehicles, pedestrians, obstacles, and road boundaries as virtual potential energy imposing constraints on the vehicle. Risk levels are quantified through potential energy intensity, field gradients, or comprehensive energy values \cite{lyu2024risk}, \cite{ma2023real}. Building upon this foundation, the academic community has further proposed unified frameworks such as driving safety fields and risk fields \cite{wang2016driving}, attempting to couple kinetic energy, potential energy, and behavioral interaction factors through a unified risk function. Some studies have also introduced probabilistic intent fields to simulate random lane-changing behaviors in mixed traffic flows \cite{huang2020probabilistic}, or integrated risk fields with control strategies for application at signalized intersections \cite{xu2023driving}.

Compared to SSMs, potential field methods provide stronger physical interpretability and support continuous risk representation. However, their practical deployment still faces two major bottlenecks. First, high precision potential fields often require frame by frame or target by target coupled computations, or iterative derivations. When applied as post processing tools to scan large scale open loop data, the computational overhead becomes prohibitive, making it difficult to satisfy high throughput screening requirements. Second, many potential field formulations remain deterministic and quasi static, and therefore cannot adequately capture directional anisotropy, multi agent long term uncertainty, or rare long tail scenarios. As a result, their effectiveness degrades in dense interaction settings and under extreme operating conditions.

\subsection{Trajectory Prediction}
As the cornerstone of autonomous driving risk assessment, trajectory prediction aims to infer the future motion path of a target vehicle within a finite time window by jointly modeling its historical motion state and contextual road environment information. Examining the evolution of this technology, existing research can be broadly categorized into three paradigms: physics-based models, classical machine learning, and deep learning.

Physics based methods center on kinematic or dynamic equations. This category encompasses linear models such as Constant Velocity (CV) and Constant Acceleration (CA), as well as families of curve models like CTRV, CTRA, and CSAV that incorporate yaw rate and steering constraints \cite{schubert2008comparison}, \cite{barth2008will}. In practical applications, physical models are often integrated with Kalman filtering and V2V measurement technologies \cite{lytrivis2008cooperative}, \cite{zhang2017method} to optimize state estimation accuracy. They are widely used for predicting distant vehicle trajectories and collision warnings. Their primary advantages lie in clear physical interpretability and low computational overhead. However, constrained by simplistic motion assumptions, they struggle to effectively capture complex multi-agent strategic coupling interactions and abrupt changes in driving intent on urban roads.

Classic machine learning methods focus on probabilistic inference through statistical modeling of historical sequences. Representative work includes generating interaction aware probabilistic predictions using dynamic Bayesian networks. \cite{schulz2019learning}, and employing Markov and hidden Markov models \cite{li2022autonomous}, \cite{qiao2014self} to describe action-state evolution while further extending adaptive parameter selection mechanisms. However, such approaches heavily rely on manually designed feature engineering and finite state space assumptions, limiting their ability to represent complex dynamic scenarios.

In recent years, deep learning has become the mainstream approach in this field due to its powerful feature extraction capabilities. Early studies predominantly employed RNNs, LSTMs, and their variants to capture long term temporal dependencies and fuse spatial interaction information through gating mechanisms \cite{dai2019modeling}, \cite{kim2017probabilistic}; Subsequently, graph neural networks \cite{xu2022adaptive} explicitly modeled multi-agent interactions through message passing mechanisms. Combined with hierarchical graph networks \cite{gao2020vectornet} for vectorized high precision maps, they further enhanced the representation of scene topology. Generative frameworks are widely employed to characterize future multimodal distributions. Transformers and their attention mechanisms substantially enhance the capacity to model long range interactions by capturing global dependencies \cite{zhu2022spatio}. Accordingly, many studies adopt proposal refinement pipelines or homogeneous encoder decoder architectures. Meanwhile, notable efficiency gains have been achieved through agent centric representations, spatiotemporal factorization, and cross attention based fusion strategies \cite{wang2023prophnet}.

Despite continuous improvements in prediction accuracy, most methods still prioritize mean displacement error as the primary objective. This tendency favors high frequency normal patterns, leading to smoothed predictions in long tail hazardous scenarios and diminishing sensitivity to critical conflicts. Consequently, research has begun integrating risk labels, safety constraints, or interpretable priors with prediction networks to jointly model high-value hazardous scenarios. This approach aims to improve the identification of high interaction segments while preserving throughput.

\section{Methodology}
In this study, we propose a high speed screening framework for identifying potentially hazardous scenarios in autonomous driving testing. The framework prioritizes the discovery and ranking of long tail high risk segments from large scale open loop naturalistic driving data with low computational overhead. It consists of three stages. (1) Comprehensive risk ground truth construction: by integrating an enhanced Driver Risk Field with a dynamic cost model, the framework computes continuous spatiotemporal risk along real trajectories, which serves as an interpretable supervision signal during training. (2) Risk aware prediction model: an encoder decoder architecture encodes multi agent historical states and high definition map context. The decoder jointly predicts future trajectories and risk sequences, while risk trajectory cross attention implements an implicit input, explicit output risk modeling mechanism, improving sensitivity to risk evolution while preserving reasonable trajectory prediction performance. (3) Screening phase: given only fixed length historical observations, the model outputs future risk sequences in a single forward pass and aggregates them into scene level risk scores, enabling large scale risk prioritization and candidate set extraction.
\subsection{Problem Formulation}
The primary objective of this paper is to enable autonomous, high efficiency screening of hazardous scenarios from large scale open loop naturalistic driving data, thereby supporting subsequent testing and validation of autonomous driving systems. The overall problem setting is illustrated in Fig. \ref{fig1}.

In summary, the traffic scenario defined in this paper consists of \textit{T} time steps in the historical period and \textit{F} time steps in the future period, encompassing a total of \textit{N} agents within the scenario. Historical observation sequences can be represented as$\{f_{T+1},f_{T+2},...f_0\}$, among which:
\begin{equation}
f_t=\{a_t^{1:N},M\},t=T+1,...,0
\end{equation}

In the formula, $a_t^n$denotes the historical state vector of the nth agent at time step \textit{t}, including position, velocity, heading angle and other information. \textit{M} represents high precision map data and road topology information. The core objective of the model is to output the risk of agent \textit{n} over the next \textit{F} time steps under mode \textit{k}:
\begin{equation}
\widehat{R}_F^{n,k}=\{\widehat{r}_{1,n,k},\widehat{r}_{2,n,k},...\widehat{r}_{F,n,k}\},n\in[1,N],k\in[1,K]
\end{equation}

The auxiliary prediction target outputs the trajectory sequence of agent \textit{n} over the next \textit{F} time steps under pattern \textit{k}:
\begin{equation}
\widehat{Y}_F^{n,k}=\{\widehat{y}_{1,n,k},\widehat{y}_{2,n,k},\widehat{y}_{F,n,k}\},n\in[1,N],k\in[1,K]
\end{equation}

In summary, the overall prediction target can be expressed as:
\begin{equation}P=\{\widehat{R}_F^{n,k},\widehat{Y}_F^{n,k}\},n\in[1,N],k\in[1,K]\end{equation}

\subsection{Risk Perception Trajectory Prediction Model}
\begin{figure*}[!t]
\centering
\includegraphics[width=1.0\textwidth]{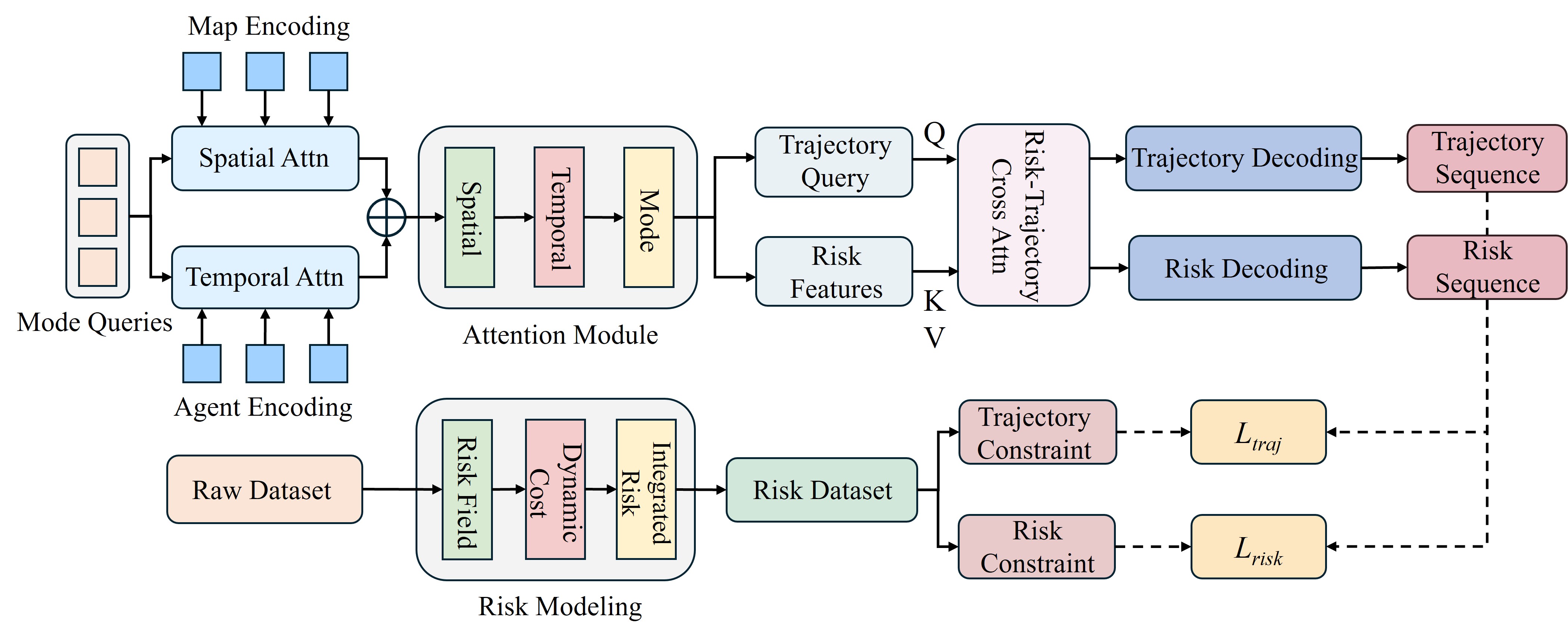}
\caption{Structure of the Risk Perception Prediction Model. An encoder decodes unified representations from multi agent motion histories and high definition map semantics, while the decoder jointly predicts future trajectories and risk sequences, treating risk prediction as the primary objective and trajectory prediction as an auxiliary task.}
\label{fig2}
\end{figure*}

As shown in Fig. \ref{fig2}, the proposed risk perception prediction model follows the mainstream encoder–decoder paradigm. The encoder extracts deep spatiotemporal representations from multi-agent state histories and high definition map semantics in traffic scenes. The decoder then jointly predicts future trajectories and potential risk sequences.

Unlike conventional prediction frameworks, the key contribution of this work is to treat trajectory prediction as an auxiliary objective while prioritizing risk prediction as the primary task. To this end, we introduce a risk–trajectory cross-attention module to strengthen risk perception. The resulting architecture adheres to an “implicit input-explicit output” risk modeling scheme: risk signals are not provided to the encoder as inputs, but are instead leveraged as supervisory targets in the decoding stage. Meanwhile, the asymmetric risk–trajectory cross-attention encourages the trajectory branch to actively attend to risk aware context, thereby improving the risk sensitivity of trajectory representations and enhancing risk perception. Through multitask learning, the model captures fine-grained spatiotemporal interactions and risk evolution during training.

\subsubsection{\textit{\textbf{Spatiotemporal Feature Encoding}}}
The primary function of the encoder is to transform dynamic and heterogeneous traffic scene information into a structured and unified feature representation. This representation provides high dimensional semantic context for the downstream decoder to jointly predict future trajectories and risk sequences. To enable efficient spatiotemporal feature extraction, the encoder leverages a two stage design that first constructs initial embeddings and then performs multi stage attention based refinement \cite{tang2024hpnet}.

First, to efficiently characterize complex traffic environments, the model leverages a vectorized representation scheme. The historical motion of each traffic participant is encoded as a state sequence that captures key kinematic attributes, including position, velocity, and orientation. In parallel, static elements from high definition maps, including lane center lines, road boundaries, and stop lines, are parsed into a set of vectorized polylines. Two multi layer perception modules then project both agent level and map level inputs into a high dimensional feature space.
\begin{equation}H_a^{t,n}=\mathrm{MLP}(a_t^n)\end{equation}
\begin{equation}H_m=\mathrm{MLP}(M)\end{equation}

Additionally, to explicitly characterize the interactive geometric relationships between agents and map elements, as well as among different agents, we introduce a relative spatiotemporal position encoding. For any element i and its domain element j, their relative positional relationship is constructed using their absolute positions $(p_i^t,\theta_i^t,t)$ and $(p_j^s,\theta_j^s,s)$ , respectively. 
\begin{equation}
\begin{aligned}
\mathrm{d}_{j\to i}^{s\to t} = \big(
&\| \mathbf{p}_s^j - \mathbf{p}_t^i \|_2, \mathrm{atan2}(\Delta y, \Delta x) - \theta_t^i, \\
&\theta_s^j - \theta_t^i,  s - t \big)
\end{aligned}
\label{eq:distance}
\end{equation}
Where$(\Delta y,\Delta x)=\mathrm{p}_i^t -\mathrm{p}_j^s$.This relative positional relationship characterizes the relative distance, relative azimuth angle relative to i's orientation, relative heading difference, and relative time interval, thereby encoding both spatial and temporal relationships within a unified representation. Subsequently, relative position embeddings are obtained via a multilayer perceptron:
\begin{equation}H_e=\mathrm{MLP}(d_{j\to i}^{s\to t})\end{equation}

After the scene is initially encoded, the model generates an initial prediction embedding $F_{t,n,k}^i$ for each future action modality \textit{k} of every agent \textit{n} through a spatiotemporal attention network. Specifically, this achieves preliminary context fusion by processing information flows across both temporal and spatial dimensions in parallel:
\begin{equation}
\begin{aligned}
F_{t,n,k}^i &= \text{TemporalAttn}\big(Q_{t,n,k}, [H_a^{t,n}, H_e], [H_a^{t,n}, H_e]\big) \\
&\quad + \text{SpatialAttn}\big(Q_{t,n,k}, [H_m, H_e], [H_m, H_e]\big)
\end{aligned}
\end{equation}
In the equation, $Q_{t,n,k}$ represents the pattern query vector, which characterizes different future behavioral patterns. It subsequently serves as a decoding anchor to extract information related to this pattern from the spatiotemporal context. To further enhance contextual feature representation, the model introduces a cascaded context refinement module based on the initial prediction embedding. This module operates within a unified attention framework, sequentially refining features across three dimensions: spatial interaction, temporal evolution, and behavioral intent. The process is expressed as follows:
\begin{equation}F_{t,n,k}^f=(\mathrm{MAttn}^\circ\mathrm{HAttn}^\circ\mathrm{AAttn})(F_{t,n,k}^i)\end{equation}
Where $\mathrm{o}$ denotes sequential stacking. Spatial interaction attention captures the intentions and impacts of surrounding traffic participants; temporal attention correlates current predictions with historical ones, ensuring trajectory smoothness and logical consistency over time; pattern attention enhances the distinctiveness between different predicted trajectories, suppressing unreasonable behavioral assumptions. Through this refinement process, the encoder ultimately outputs a high dimensional context aware feature embedding , which deeply integrates spatial interactions, temporal evolution, and behavioral intent features. This provides the decoder with rich representational information containing risk priors for simultaneously performing risk prediction and trajectory prediction.

\subsubsection{\textit{\textbf{Decoding and Risk Prediction}}}
After the encoder produces context aware feature embeddings, the decoder jointly decodes these representations into future trajectory sequences and risk sequences to support autonomous screening of hazardous scenarios. To this end, we design a decoding architecture that incorporates risk trajectory cross attention and adheres to an implicit input, explicit output risk modeling mechanism. Risk signals are not provided to the encoder as input features; instead, they are learned during decoding under ground truth supervision and explicitly output as a risk sequence. This design enables joint modeling of risk prediction and trajectory generation within a unified framework.

The core of the decoder is a risk-trajectory cross-attention module. Agent features $F_{t,n,k}^f$ from the encoder first undergo two independent linear transformations, generating the trajectory query vector $Q_{traj}$ and the risk context $F_{risk}$ vector respectively. Subsequently, these two vectors enter a multi-layer cross-attention module for deep interaction. Specifically, we uses the trajectory query vector as the query vector for the attention module, while the risk context vector undergoes additional processing to serve as both the key $K_{risk}$ and value $V_{risk}$ This asymmetric design compels the trajectory prediction stream to actively extract risk information most relevant to the current trajectory intent from the risk context. Through multilayer stacked cross-attention and residual connections, trajectory features are iteratively refined by risk information, represented as:
\begin{equation}F_{t,n,k}^c=\mathrm{CrossAttn}(Q_{traj},K_{risk},V_{risk})\end{equation}
To incorporate risk information while minimizing its disruptive impact on the accuracy of original trajectory predictions, we further employ a feature concatenation decoding strategy. 
\begin{figure*}[!t]
\centering
\includegraphics[width=1.0\textwidth]{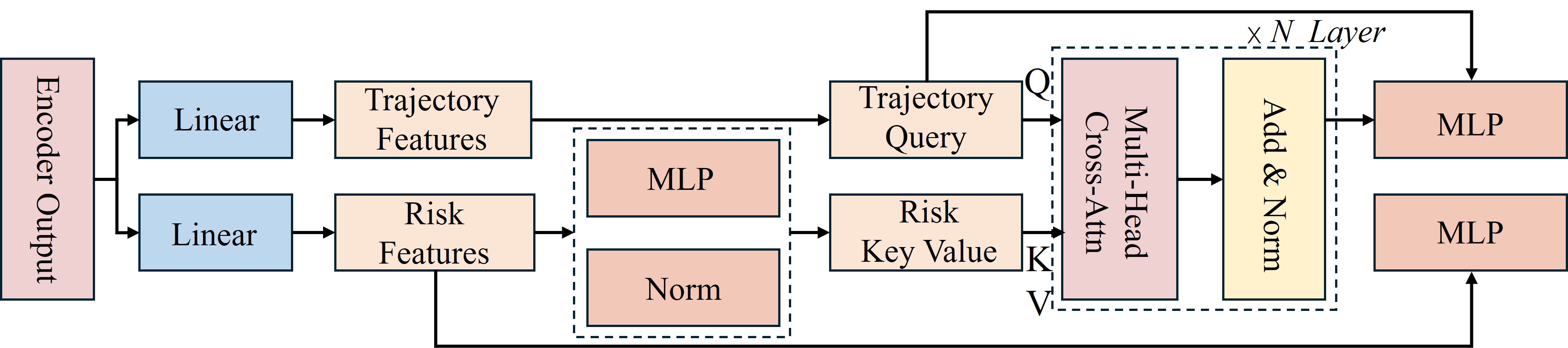}
\caption{Decoder Structure. The risk branch outputs future risk sequences under supervision, and an asymmetric cross attention module forces the trajectory branch to attend to risk context, improving risk sensitivity of trajectory representations while preserving trajectory fidelity for downstream screening.}
\label{fig3}
\end{figure*}
This approach concatenates the risk perspective features output by the cross-attention module with the original trajectory query vector prior to entering the attention module, forming a 2d composite feature. This composite feature is then fed into a dedicated MLP decoder head to regress the trajectory sequence $\widehat{Y}_F^{n,k}=\{\widehat{y}_{1,n,k},\widehat{y}_{2,n,k},...\widehat{y}_{F,n,k}\},n\in[1,N],k\in[1,K]$, where $\hat{y}_{t,n,k}=\mathrm{MLP}_{traj}([Q_{traj};F_{t,n,k}^c])$. Simultaneously, retain an independent MLP decoder to directly regress the risk sequence from the risk context features $\widehat{R}_{F}^{n,k}=\{\widehat{r}_{1,n,k},\widehat{r}_{2,n,k},...\widehat{r}_{F,n,k}\},n\in[1,N],k\in[1,K]$, where $\hat{r}_{t,n,k}=\mathrm{MLP}_{risk}(F_{risk})$. Furthermore, we extends vehicle risk prediction to scenario risk metrics. By aggregating the risk sequences of all vehicles in scenario C, we obtain 
\begin{equation}R(C)=\mathrm{Agg}(\{\widehat{R}_F^{n,k}\})\end{equation}
$\mathrm{Agg}(\cdot)$ denotes the aggregation operator. By sorting the magnitude of $R(C)$, it enables rapid construction of a risk-prioritized test set, thereby efficiently screening potential hazardous scenarios within large scale test data.

\subsubsection{\textit{\textbf{Training Objective}}}
To achieve joint modeling of trajectory prediction and risk prediction, we design a multitask training objective. The overall loss function consists of two components: trajectory prediction loss and risk prediction loss:
\begin{equation}L=\lambda_{traj}L_{traj}+\lambda_{risk}L_{risk}\end{equation}
where $\lambda_{traj}$ and $\lambda_{risk}$ are weight coefficients.

To enhance the multimodal capability of trajectory prediction, we adopts a winner-takes-all strategy \cite{lee2016stochastic}. Among the multiple candidate trajectories generated by the model, the one with the smallest endpoint error relative to the true trajectory is selected as the supervised target. Assuming the endpoint of the true trajectory $G_{F'}^{n}$, the optimal mode index is
\begin{equation}
k^* = \arg \min_{k \in [1,K]} \sum_{n=1}^N \|\widehat{Y}_{F'}^{n,k} - G_{F'}^{n}\|
\end{equation}
where ${Y}_{F'}^{n,k}$ denotes the predicted endpoint position of the nth agent under mode \textit{k}. Subsequently, the trajectory regression loss employs a smooth L1 loss to progressively align the predicted positions with the actual positions
\begin{equation}
L_{traj} = \frac{1}{NF} \sum_{n=1}^N \sum_{t=1}^F L_{Huber}\big(\widehat{Y}_F^{n,k^*}, G_F^n\big)
\end{equation}

The supervisory signals for risk prediction originate from actual risk values $R_F^n$ computed based on the driver risk field and dynamic cost model. We employ a smoothed L1 loss to define the risk prediction loss:
\begin{equation}
L_{risk} = \frac{1}{NF} \sum_{n=1}^N \sum_{t=1}^F L_{Huber}\big(\widehat{R}_F^{n,k^*}, R_F^n\big)
\end{equation}

Through the aforementioned multitask optimization, the model presented in this paper maintains its capability for multi-modal generation and high precision fitting in trajectory prediction while learning latent hazardous trends that evolve over time for risk prediction. The trajectory and risk tasks complement each other during training: trajectory prediction provides kinematic priors for risk modeling, while risk supervision imposes constraints on trajectory generation to prevent underestimation of hazardous scenarios. Consequently, the model acquires kinematic prior knowledge through trajectory prediction, thereby providing a reliable foundation for the core risk prediction task and ultimately enabling efficient screening of hazardous scenarios.

\subsection{Risk Modeling}
In autonomous driving testing and validation, risk quantification is a central element for identifying and assessing potentially hazardous scenarios. Common proxy safety metrics in the literature include time to collision (TTC) and time headway (THW). TTC estimates the remaining time until a collision between two vehicles under their current relative motion \cite{gettman2003surrogate}, whereas THW characterizes the longitudinal safety margin in car following. Owing to their low computational cost and high interpretability, these metrics have been widely adopted in traffic safety analysis and car following models. Nevertheless, they exhibit notable limitations. First, TTC and THW rely on simplified two vehicle kinematic assumptions and therefore cannot adequately capture interactions among multiple agents, geometric constraints induced by lane topology, or driver perception. Second, their strong sensitivity to threshold settings often yields discontinuous, step like risk estimates, which hinders continuous and fine grained characterization of risk evolution in space and time.

To alleviate these limitations, we adopts a risk modeling framework that is more consistent with drivers’ cognitive mechanisms. Specifically, risk is formulated as the product of two components: the probability of event occurrence and the severity of event consequences. The former is modeled via an enhanced Driver Risk Field, which captures drivers’ subjective perception of potential collision regions through a spatial probability distribution. The latter is represented by a dynamic cost model that quantifies interaction costs from the perspective of physical dynamics. By integrating these two components, the proposed framework yields a unified, continuous, and physically grounded risk metric, providing a robust basis for subsequent risk prediction and hazardous scenario screening.

\subsubsection{\textbf{Risk Distribution}}
The Driver Risk Field characterizes drivers’ subjective perception of potential collision zones during driving. The core idea is to model risk as a potential field over the vehicle’s future reachable space, thereby capturing drivers’ sensitivity to prospective hazards in the forward environment. Building on the annular Gaussian model \cite{kolekar2020human}, we propose an improved formulation that is better suited to low speed scenarios and complex lane geometries.

We define the risk field distribution at a future forward distance s as follows:
\begin{equation}DRF(s,t)=a(s)\cdot\exp\left(-\frac{t^2}{2\sigma^2(s,\delta)}\right)\end{equation}
Where $a(s)$ denotes the height function of the risk field, $\sigma(s,\delta)$ represents the lateral spread width of the risk field, and $\delta$ is the steering angle. The height function is defined as:
\begin{equation}
a(s) = 
\begin{cases}
H_0 \frac{(s_{max}(v) - s_{eff})^2}{(s_{max}(v) - s_{min})^2}, & 0 < s < s_{max}(v) \\
0, & s \ge s_{max}(v)
\end{cases}
\end{equation}
The minimum lookahead distance $S_{min}$ mitigates numerical singularities. $H_0$ denotes an amplitude parameter that bounds the peak intensity of single point risk within a fixed range, thereby preventing local overestimation. The maximum lookahead distance is formulated in a speed adaptive manner to reflect the physical intuition that higher speeds correspond to larger potential hazard zones.
\begin{equation}s_{max}(\nu)=d_s\tilde{\nu}^{\gamma_s}\end{equation}
\begin{equation}\tilde{\nu}=\nu_0+\frac{1}{k}\mathrm{ln}(1+e^{k(\nu-\nu_0)})\end{equation}
where $d_s$, $\gamma_s$, $v_0$ and $k$ are tuning parameters, $\tilde{\nu}$ ensures continuity across the low-to-high speed range, and increasing speed extends the longitudinal coverage of the risk field. The width function is defined as:
\begin{equation}\sigma(s,\delta)=\beta_ws_\mathrm{eff}+w_0+k_i|\delta|s_\mathrm{eff}\end{equation}
$\beta_w$ represents the linear expansion controlling width, $w_0$ denotes the base width, and $k_i$ signifies the asymmetric expansion distinguishing the inside and outside of turns.
\begin{figure}[!t]
\centering
\includegraphics[width=0.45\textwidth]{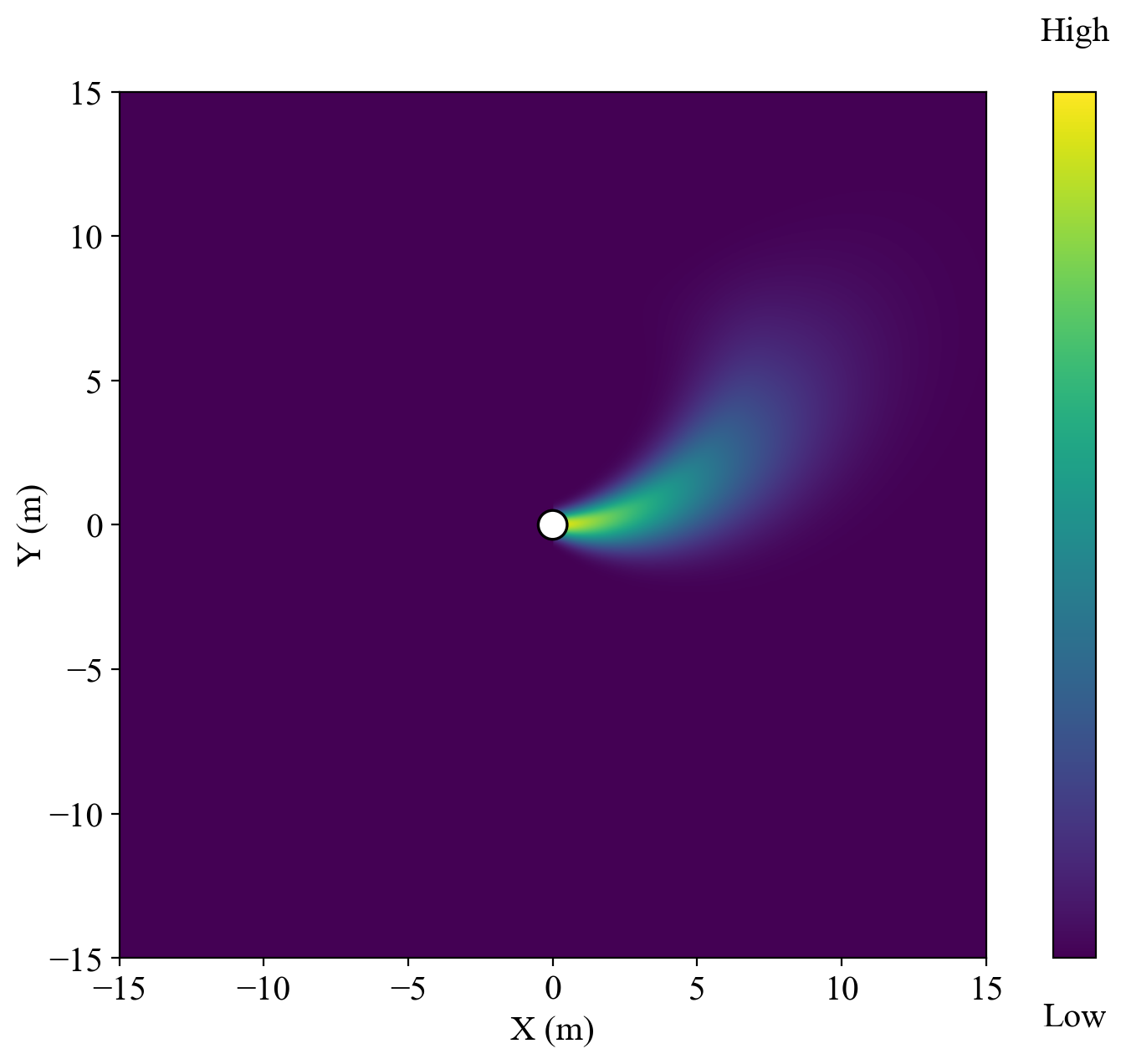}
\caption{Risk Distribution Modeling. The proposed comprehensive risk is constructed by coupling an enhanced driver risk field that models the probability of entering potential collision zones with a dynamic cost model that quantifies consequence severity under vehicle dynamics and interactions.}
\label{fig4}
\end{figure}

\subsubsection{\textbf{Risk Cost}}
In real world traffic scenarios, relying solely on the spatial distribution of the Driver Risk Field is insufficient to fully characterize the physical severity of collision outcomes. Risk between vehicles is determined not only by spatial proximity but also by dynamic factors such as speed, acceleration, and vehicle mass. In particular, collision consequences can escalate rapidly during high speed merging maneuvers or under large relative velocities. To account for these effects, we introduce a dynamic cost modeling module. By jointly modeling individual dynamics and pairwise interaction dynamics, the proposed module constructs a risk cost field that captures the temporal evolution of collision severity. Specifically, it computes interaction costs for all agent pairs at each time step, and then incorporates geometric boundary constraints, spatial diffusion, and multi-agent aggregation to derive an overall cost map for the target vehicle. The modeling pipeline is organized into four steps, as detailed below.

\textbf{a) Cost Function Definition}:
For any two agents, the interaction cost function is defined as follows:
\begin{equation}
\begin{aligned}
C_{ij} &= w_b C_{base} + w_a \ln\!\left(1 + \frac{1}{2} M \|v_j\|^2\right) \\
       &+ w_r \ln\!\left(1 + \frac{1}{4} M \|v_j - v_i\|^2\right)
\end{aligned}
\end{equation}

The first term $C_{base}$ represents the baseline cost, ensuring that the risk value remains greater than zero even at low speeds or when stationary, thereby preventing risk gaps. The second term corresponds to the absolute kinetic energy component, which characterizes the hazard induced by an obstacle’s own motion. Higher speeds imply greater potential risk. The third term corresponds to the relative kinetic energy component, which captures the collision kinetic energy arising from relative vehicle speeds. Larger speed differences therefore lead to higher risk costs. $w_b$ is the dimensionless weighting factor, while $w_a$ and $w_r$ are kinetic energy risk conversion coefficients, both measured in RiskUnit $\cdot \mathrm{J}^{-1}$.

\textbf{b) Directional Bounding Box Geometric Constraints}:In real world traffic scenarios, vehicles are not idealized points but possess distinct geometric shapes. To more accurately delineate vehicle boundaries, we employ Oriented Bounding Boxes to compute the closest distance between the target vehicle's position and the OBB of obstacle j:
\begin{equation}
d(x,y;j) =
\begin{cases}
0, & (x,y) \in \mathrm{OBB}_j \\
\mathrm{dist}\left((x,y), \partial\mathrm{OBB}_j\right), & (x,y) \notin \mathrm{OBB}_j
\end{cases}
\end{equation}

When a grid point lies within an obstacle's bounding box, the distance is defined as zero; if it lies outside the bounding box, the shortest Euclidean distance to the bounding box boundary is calculated. This design not only ensures geometric accuracy in spatial modeling but also effectively distinguishes between different risk zones around the vehicle.

\textbf{c) Gaussian Kernel Diffusion Smoothing}:
Relying solely on geometric boundaries can cause risk distributions to exhibit hard edges, hindering gradient propagation in continuous scenarios. Therefore, we apply Gaussian-kernel diffusion to smooth the cost, so that the risk can diffuse outward with distance, with progressively smaller contributions as the distance increases.
\begin{equation}w(x,y;j)=\exp\left(-\frac{1}{2}{\left(\frac{d(x,y;j)}{\sigma}\right)^2}\right)\end{equation}

Thus, the cost contribution of a single obstacle j at position $(x, y)$ is obtained.
\begin{equation}\mathrm{cost}\left(x,y;j\right)=C_{ij}\cdot w(x,y;j)\end{equation}

Through this process, the originally discrete geometric boundaries are expanded into a cost distribution with continuous attenuation characteristics, enabling a more realistic simulation of a driver's perception of nearby hazards.

\textbf{d) Multi-Obstacle Cost Aggregation}:
In complex multi vehicle interaction scenarios, the risk to the target vehicle arises not from a single obstacle but from the combined actions of multiple agents. The final cost map is formed by aggregating the cost values of all interacting objects, weighted by their respective contributions.
\begin{equation}\mathrm{CostMap}_i(x,y)=\sum C_{ij}\cdot w(x,y;j)\end{equation}

Dynamic cost modeling not only incorporates kinetic factors at the numerical level but also integrates vehicle OBB representation and spatial diffusion mechanisms at the geometric level, providing a stable and interpretable severity metric for subsequent risk fusion.

\subsubsection{\textbf{Integrated Risk}}
After completing modeling of the risk field based on driver perception and dynamic cost modeling based on dynamic properties, we further integrate the two approaches to simultaneously capture spatial perception risks and dynamic interaction risks in traffic scenarios. The former primarily depicts the risk distribution within the vehicle's surrounding space as perceived subjectively by the driver, excelling in simulating the driver's forward viewing characteristics and potential collision probability. The latter, grounded in physical dynamics, quantifies the impact of speed, mass, and relative motion states on the severity of potential hazards.

Table \ref{Trajectory Prediction Accuracy.} presents quantitative results for multimodal prediction. The model achieves a minimum joint mean displacement error of 0.4409 meters and a minimum joint final displacement error of 1.0649 meters, demonstrating robust prediction accuracy and validating trajectory fitting.

Therefore, we propose integrating risk distribution with risk cost to construct a comprehensive risk potential function, defined as follows:
\begin{equation}R_i^n=f_{\mathrm{DRF}}\cdotp f_{\mathrm{COST}}\end{equation}

$R_i^n$ denotes the comprehensive risk of agent \textit{n} at time \textit{i}. For the target vehicle, its comprehensive risk equals the sum of all subject risks. This fusion approach integrates the spatial distribution provided by the risk field with the dynamic weights supplied by the cost model, thereby mechanistically achieving a coupled quantification of both the probability of risk occurrence and the severity of risk consequences.

\section{Experiment}
\subsection{Experimental Setup}
\subsubsection{\textbf{Experimental Data}}

The INTERACTION dataset \cite{zhan2019interaction} focuses on trajectory prediction tasks, collecting diverse and complex scenario data from urban roads. It covers typical traffic scenarios such as highway merging, signalized intersections, and roundabouts, spanning driving behaviors across multiple countries and varying traffic regulations. The dataset contains over one million vehicle and pedestrian trajectories sampled at a frequency of 10 Hz. The FLUID dataset \cite{chen2025fluid} targets high risk traffic environments, collected at urban signalized intersections with extremely high traffic conflict density. The intersections where it was collected experience approximately 2 motor vehicle conflict events per minute on average, with about 24.85\% of vehicles involved in potential conflicts during the sampling period. The number of risk events is significantly higher than in typical natural driving datasets, and it includes real world risk labels. This provides a massive amount of authentic high risk samples for evaluating a model's performance in risk identification and hazardous scenario screening. To validate the model's adaptability and transferability to new scenarios, the proposed model was trained and tested on the INTERACTION dataset, while the FLUID dataset served as the novel scenario for evaluation.

\subsubsection{\textbf{Experimental Environment}}
The development tool used for this model is Visual Studio Code. Training and experiments were conducted on a computing device configured with an AMD EPYC 9754 CPU and an NVIDIA RTX 4090D GPU, running the Ubuntu 22.04 operating system. The model was developed using Python 3.10 and PyTorch 2.1.0. The AdamW optimizer was employed with an initial learning rate of $4\times10^{-4}$, a batch size of 16, a dropout rate of 0.1, and a weight decay rate of $1\times10^{-4}$. Cosine annealing was used for learning rate decay, with a maximum training epoch of 72.

\subsection{Analysis of Trajectory Prediction Results}
\begin{figure*}[!t]
\centering
\includegraphics[
  width=1.0\textwidth,  
  keepaspectratio,      
]{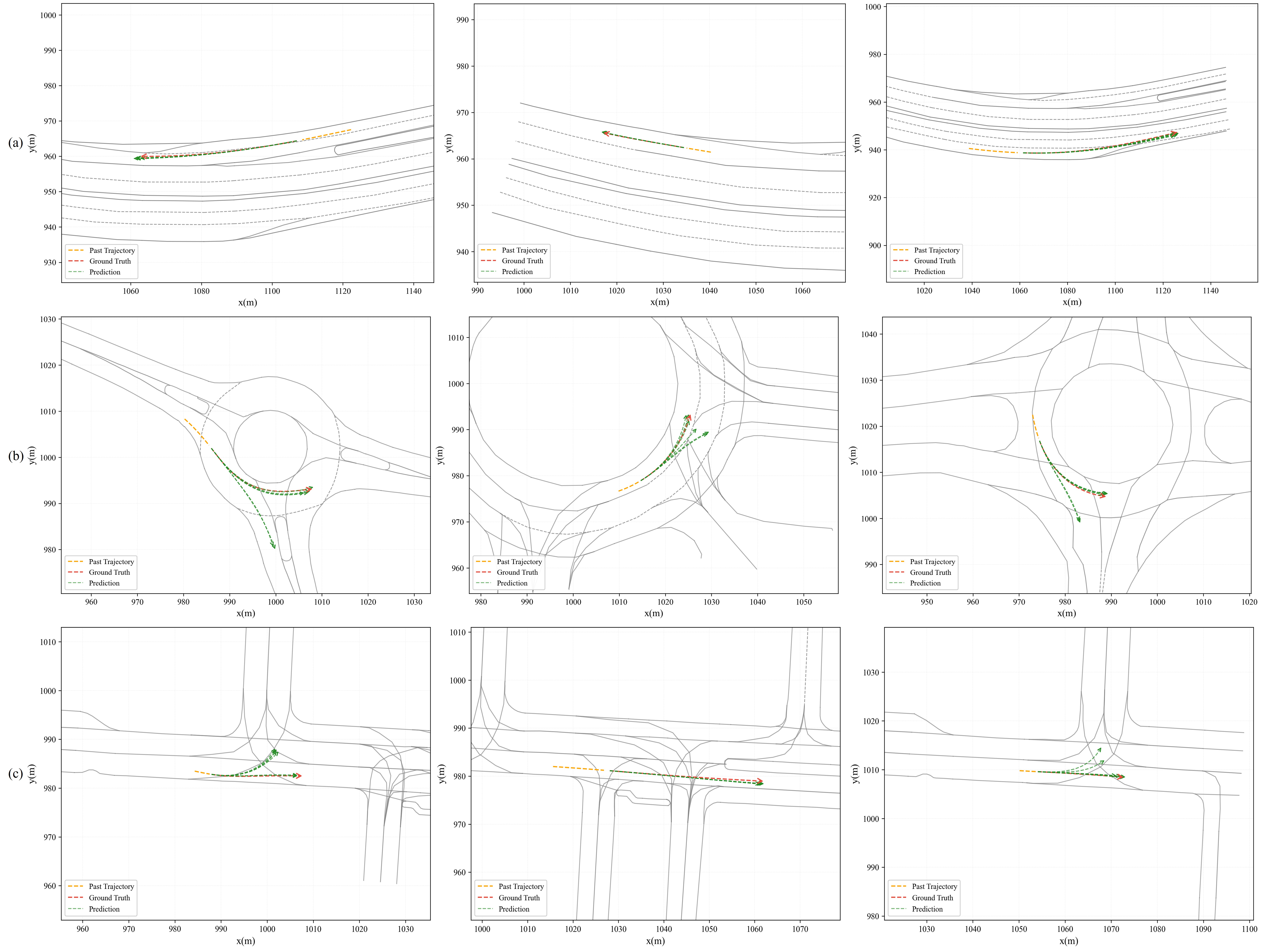}
\caption{Multimodal trajectory prediction results for representative scenarios. Qualitative trajectory predictions in merging, roundabout, and intersection scenes, where the model produces multiple plausible futures that cover the ground truth and provide reliable kinematic hypotheses for risk estimation and screening.}
\label{Multimodal trajectory prediction results}
\end{figure*}

Fig. \ref{Multimodal trajectory prediction results} presents the multimodal trajectory prediction results of the proposed model in representative scenarios. The yellow trajectory represents the historical trajectory, the green trajectory indicates the predicted trajectory, and the red trajectory denotes the actual trajectory.

The proposed model undergoes multi-modal trajectory prediction performance evaluation across various typical and complex driving scenarios. (a) In merge and lane-change scenarios, despite significant lateral displacement and longitudinal speed adjustments, the model's predicted trajectories closely align with actual trajectories and exhibit concentrated distributions. This indicates the model accurately captures vehicles' lane-change intentions and lane-keeping capabilities. (b) In roundabout scenarios, the model successfully fits the dynamic characteristics of vehicles traveling around or exiting the roundabout despite high-curvature nonlinear motion, highlighting its robustness under continuous steering. (c) At intersections, the model effectively identifies vehicles' straight-through or turning intentions. Despite high behavioral uncertainty in intersection environments, the top modalities of prediction results precisely cover actual trajectories, demonstrating the model's deep understanding of road topology and traffic interaction logic. Overall, the proposed prediction model exhibits outstanding multimodal modeling capabilities, providing high fidelity future motion hypotheses under complex conditions and offering reliable kinematic prior support for downstream risk scenario screening.
\begin{table}[htbp]
  \centering
  \captionsetup{skip=0pt}
  \caption{Trajectory Prediction Accuracy.}
  \label{Trajectory Prediction Accuracy.}
  
  \begin{tabular}{w{c}{2.0cm} w{c}{2.5cm} w{c}{2.5cm}}
    \toprule
    Metric & minJointADE & minJointFDE \\  
    \midrule
    Value  & 0.4409      & 1.0649      \\
    \bottomrule
  \end{tabular}
\end{table}

\subsection{Analysis of Risk Modeling Results}
\begin{figure}[!t]
\includegraphics[width=0.46\textwidth]{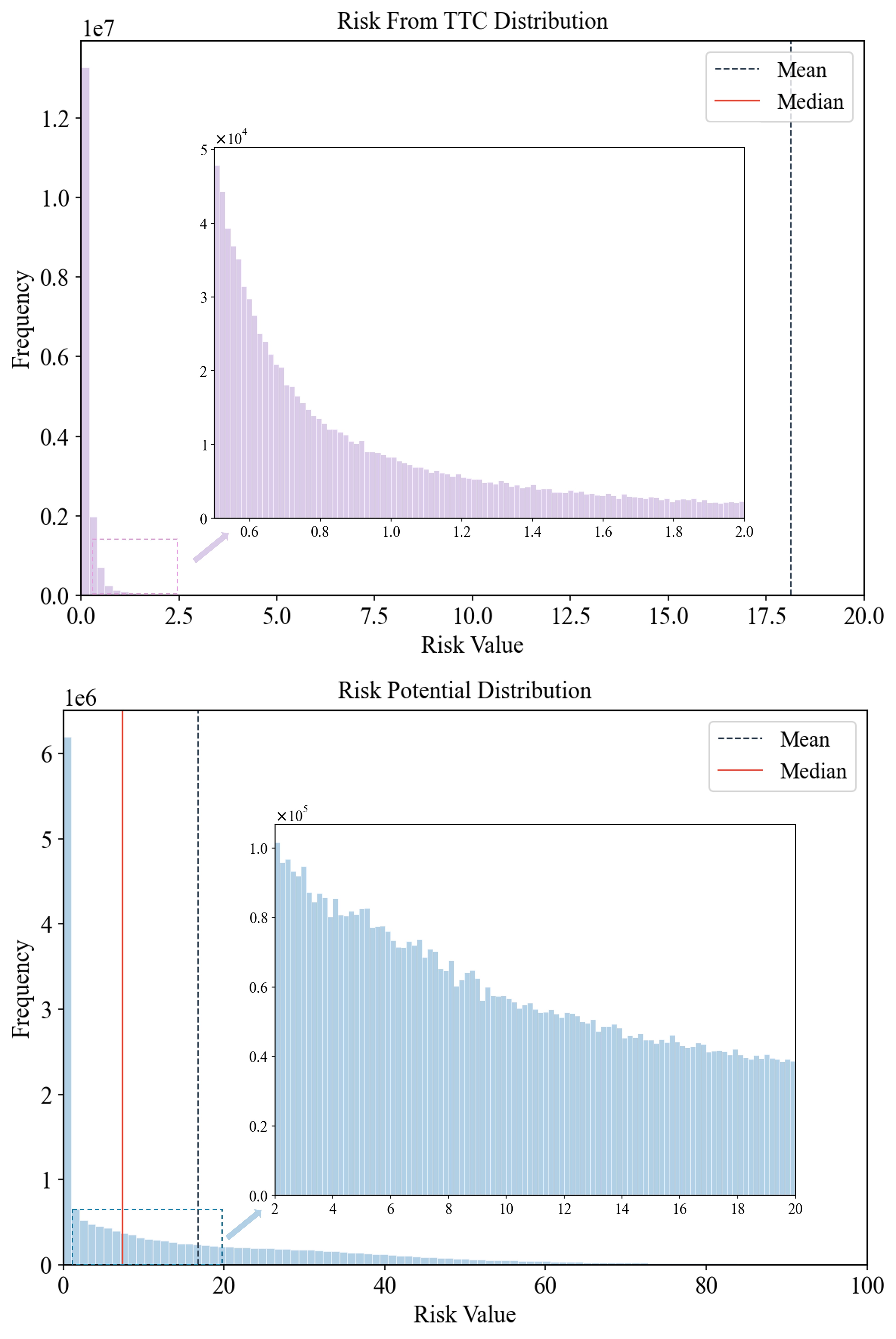}
\caption{1/TTC vs. Comprehensive Risk Distribution Chart. Histogram level comparison showing that 
1/TTC is sparse and dominated by extreme near collision events, whereas the proposed comprehensive risk assigns informative baseline risk to broader interaction contexts, supporting earlier and smoother screening signals.}
\label{1/TTC vs. Comprehensive Risk Distribution Chart}
\end{figure}

To validate the applicability of the proposed risk modeling method under varying traffic conditions, we first compares the distribution characteristics of the model-generated comprehensive risk scores with those of traditional proxy indicators. For ease of comparison, TTC is converted to 1/TTC, where higher values indicate greater risk, aligning with the trend observed in the risk modeling results. The distribution of the comprehensive risk scores and 1/TTC is illustrated in Fig. \ref{1/TTC vs. Comprehensive Risk Distribution Chart}.

\begin{table}[htbp]
  \centering
  \captionsetup{skip=0pt}
  \caption{Statistical Comparison of 1/TTC and Comprehensive Risk.}
  \begin{tabular}{w{c}{2.0cm} w{c}{2.5cm} w{c}{2.5cm}}
    \toprule
    Metric   & Risk From TTC & Risk Potential \\
    \midrule
    Mean     & 18.13         & 16.80          \\
    Variance & 57.04         & 23.24          \\
    Median   & 0.00          & 7.41           \\
    \bottomrule
  \end{tabular}
  \label{tab:risk_stats}
\end{table}
As shown in Fig. \ref{1/TTC vs. Comprehensive Risk Distribution Chart}, the distribution of 1/TTC exhibits a pronounced long-tail characteristic, with a mean of 18.13 and a median of 0. This indicates that over half of the scenarios do not present direct risk in a purely kinematic sense, while only a few extremely small TTC events significantly skew the overall mean upward. This exposes the sparsity and lag in TTC as a risk metric. In contrast, the risk generated by the proposed model has a mean of 16.81 and a median of 7.41. This indicates that the model assigns meaningful baseline risk to numerous scenarios deemed zero-risk by TTC, moving beyond reliance on imminent collision scenarios to capture broader contextual information that foreshadows potential hazards.

\begin{figure}[!t]
\includegraphics[width=0.46\textwidth]{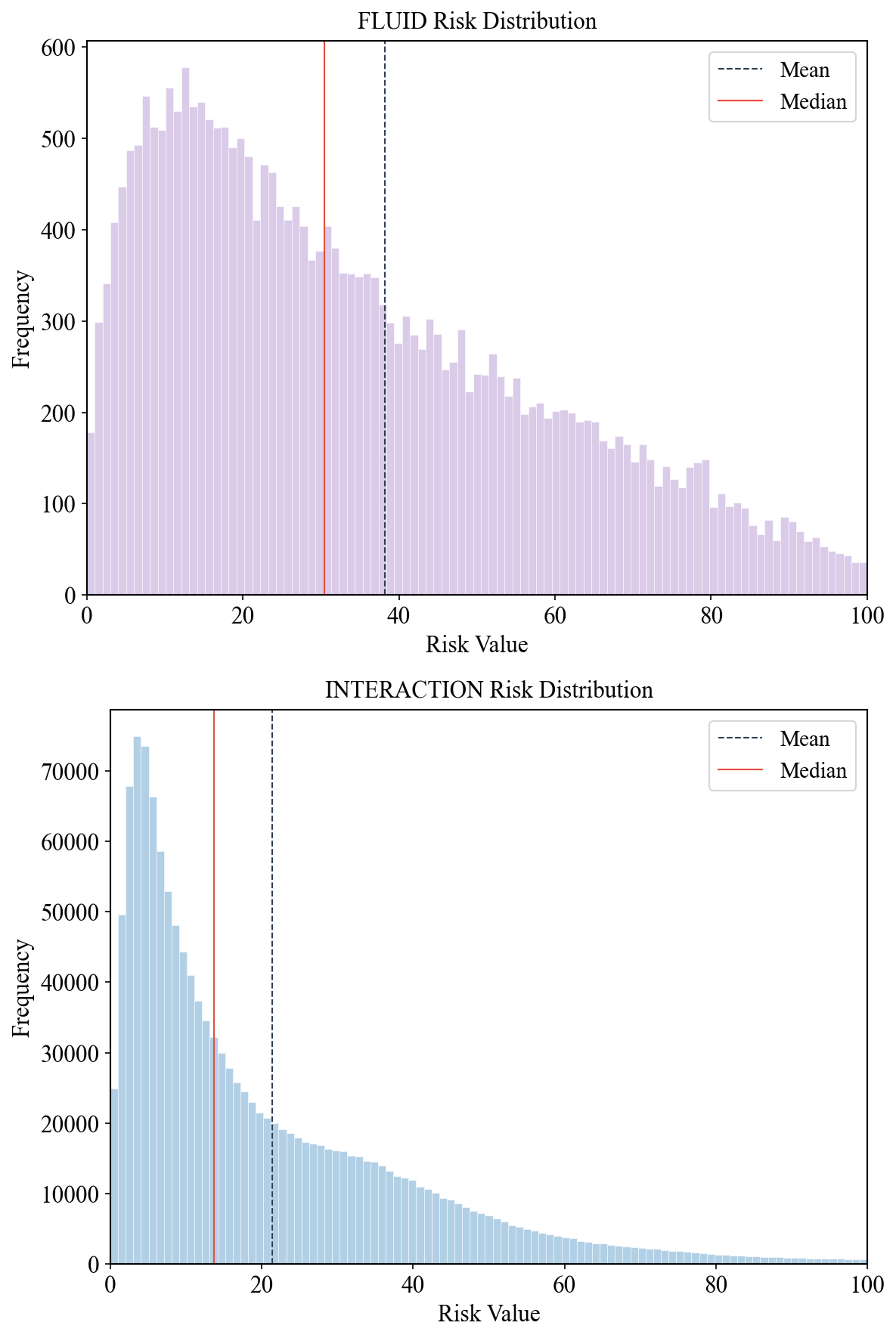}
\caption{Risk Distribution of FLUID and INTERACTION Datasets. The model outputs distinguish routine interaction data (INTERACTION) from high conflict data (FLUID) by a pronounced right shift and higher density in risky intervals, demonstrating sensitivity to different traffic environments.}
\label{Risk Distribution of FLUID and INTERACTION Datasets}
\end{figure}

Furthermore, testing on the INTERACTION and FLUID datasets enabled visualization analysis of the risk value distribution from the model outputs. As shown in Fig. \ref{Risk Distribution of FLUID and INTERACTION Datasets}, the risk distribution on the INTERACTION dataset exhibits a long tail characteristic similar to 1/TTC, consistent with the dataset's focus on routine driving and general interaction scenarios. In contrast, on the high conflict density FLUID dataset, the overall risk distribution shifts significantly to the right, with peaks concentrated in higher risk intervals and a more uniform distribution. This indicates the model's ability to effectively distinguish risk levels across different traffic environments and accurately identify the high intensity interaction risks prevalent in the FLUID dataset.

To further validate the effectiveness and accuracy of model risk assessment, we select specific interaction cases predicted by the model to visually demonstrate the dynamic evolution of risk over time.

\begin{figure*}[!t]
\includegraphics[width=1.0\textwidth]{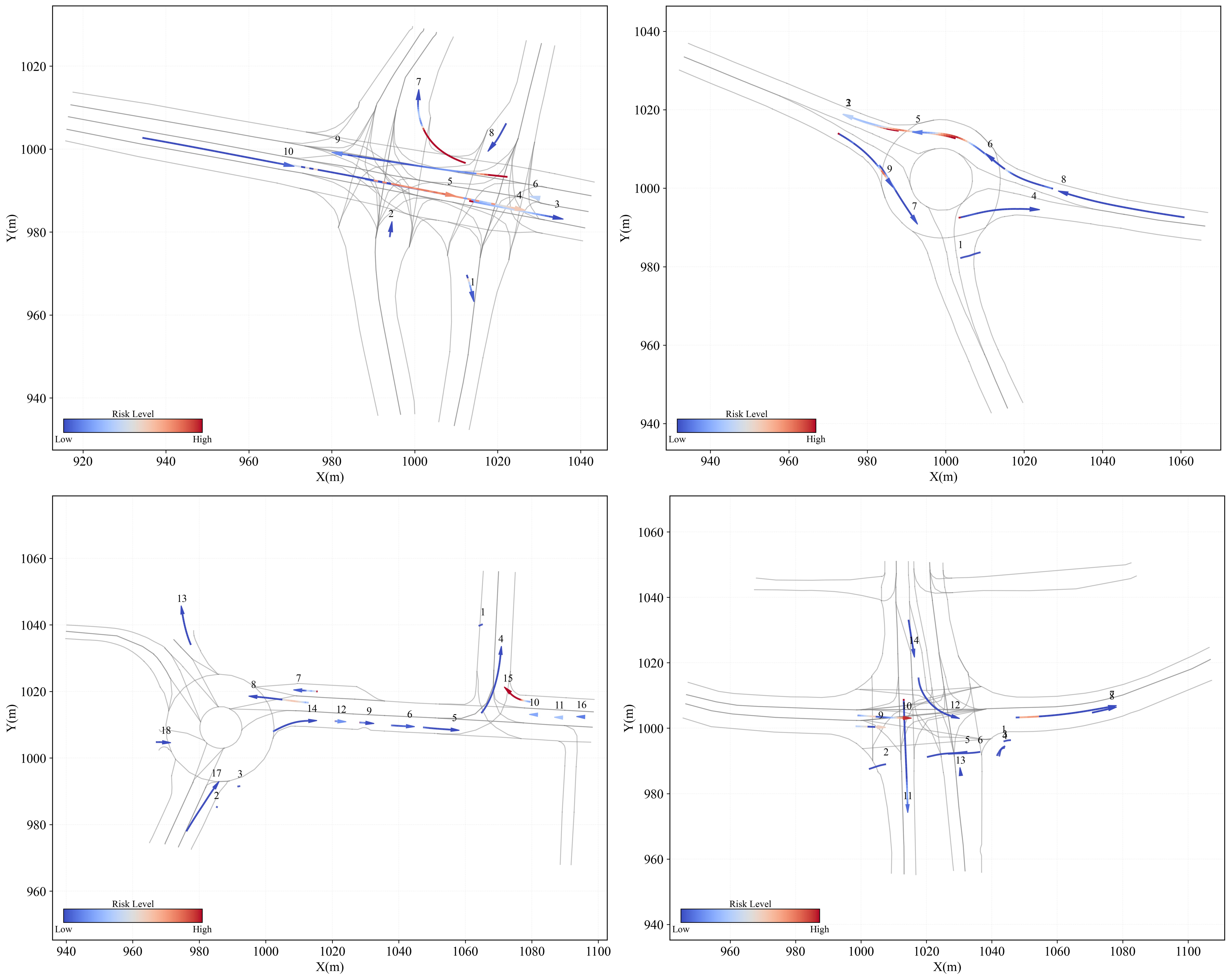}
\caption{Vehicle Trajectories and Risk Distribution in a Typical Scenario. Example visualizations across diverse scenes, illustrating localized high risk regions near conflict points and showing that the proposed risk metric can reveal the overall multi agent risk landscape beyond pairwise kinematic indicators. }
\label{Vehicle Trajectories and Risk Distribution in a Typical Scenario}
\end{figure*}

In intersection scenario (a), some left-turning and straight-through vehicles exhibit a distinct high-risk distribution within the conflict zone, with the model accurately identifying potential collision points. In roundabout scenario (b), the risk associated with entering and exiting vehicles rapidly increases at the roundabout junction, indicating the model's ability to capture hazardous trends during multi-vehicle merging. In the roundabout scenario (c), vehicles 15 and 4 exhibit significant interaction at the roundabout entrance, with highly overlapping future trajectories forming a localized high risk zone. This demonstrates the model's sensitivity in detecting vehicles at potential conflict. In the intersection scenario (d), vehicles at the convergence point are labeled as high-risk, intuitively reflecting the danger associated with vehicles in potential conflict zones. These results demonstrate that the proposed method not only characterizes individual risk evolution in single cases but also comprehensively reveals the overall risk landscape of multi-agent systems across diverse traffic scenarios. This provides robust support for identifying and testing hazardous scenarios in complex traffic environments.

\begin{figure}[!t]
\includegraphics[width=0.5\textwidth]{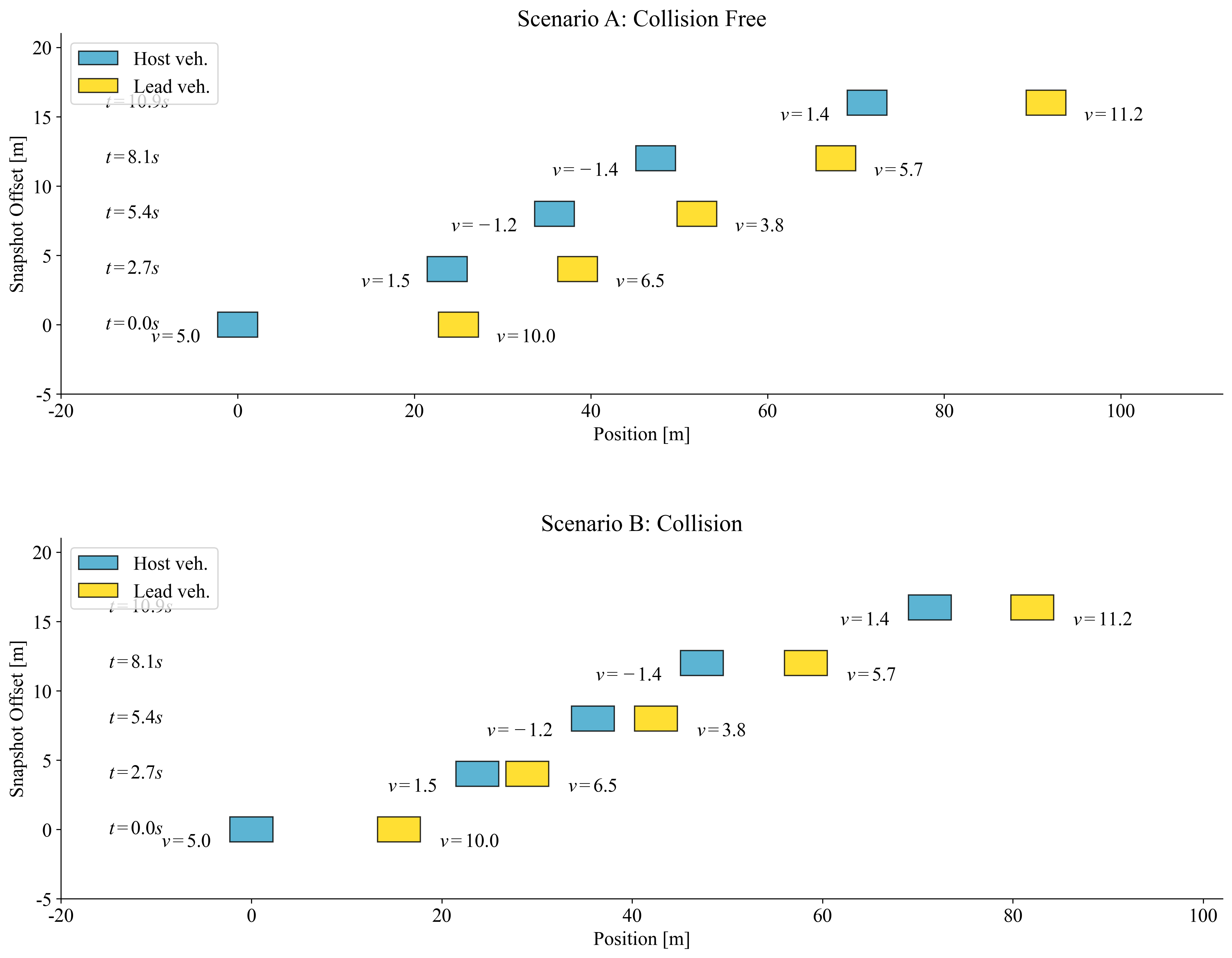}
\caption{Safe Following Scenario A and Rear-End Collision Scenario B. Two representative following cases from FLUID, used to contrast how different indicators respond to safe interactions versus critical conflict accumulation leading to collision.}
\label{Safe Following Scenario A and Rear-End Collision Scenario B}
\end{figure}

To thoroughly validate the sensitivity and validity of this risk model under extreme operating conditions, Fig. \ref{Safe Following Scenario A and Rear-End Collision Scenario B} presents a comparative analysis of two typical following interaction scenarios selected from the FLUID dataset. Scenario A represents safe following, while Scenario B depicts a high risk scenario involving a rear end collision.

\begin{figure*}[tp]
\centering
\includegraphics[width=0.95\textwidth]{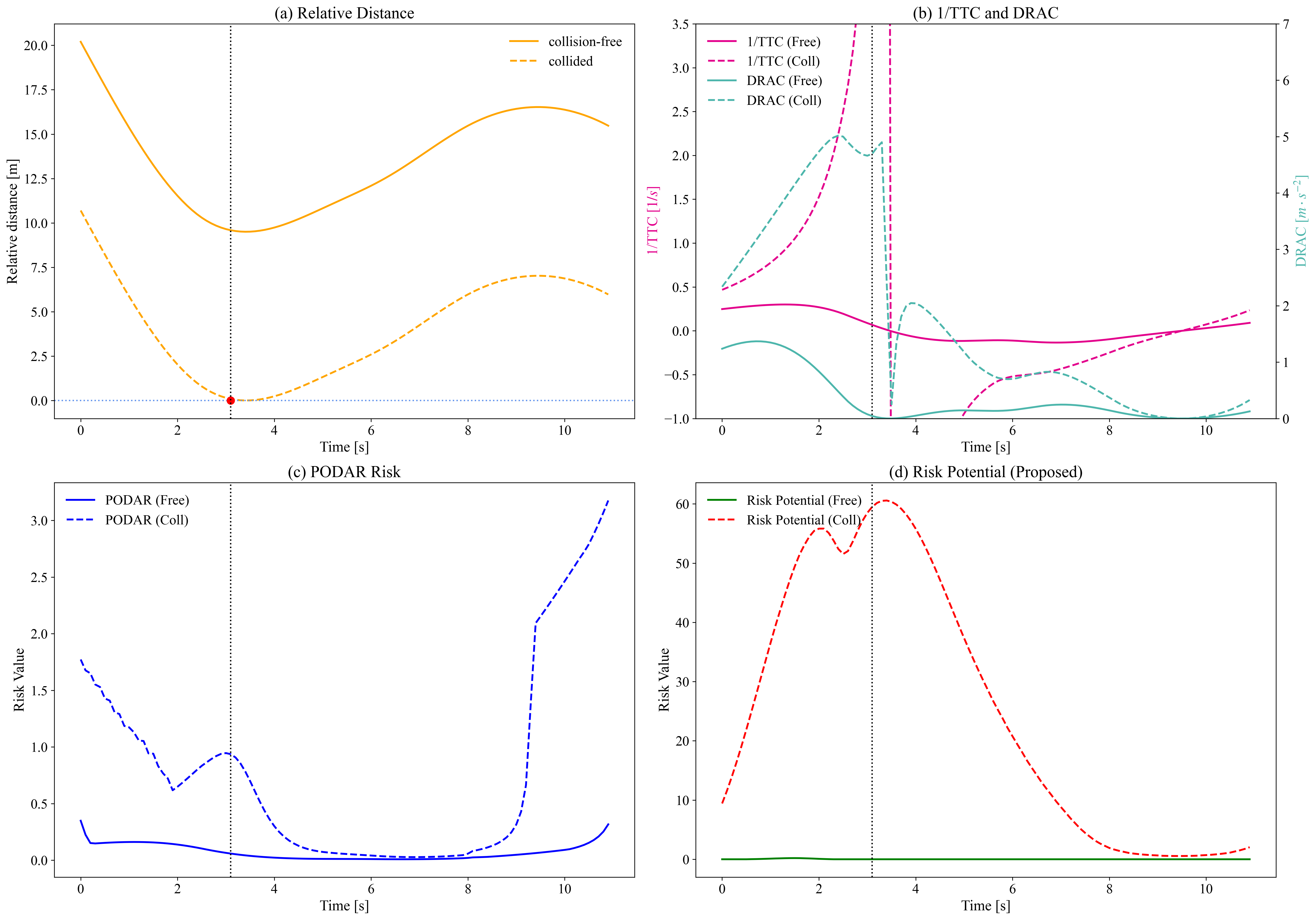}
\caption{Curves of Various Metrics Under Safe and Critical Scenarios. Temporal evolution of multiple proxy safety measures and the proposed comprehensive risk, highlighting that traditional kinematic metrics exhibit lag, sparsity, or oscillation, while the proposed risk increases continuously and peaks near the conflict moment with early warning capability.}
\label{Curves of Various Metrics Under Safe and Critical Scenarios}
\end{figure*}

As shown in the traditional indicator curve in Fig. \ref{Curves of Various Metrics Under Safe and Critical Scenarios} (b), 1/TTC exhibits significant lag and sparsity characteristics. In Scenario B, its value only exhibits a step-like surge in the extremely brief period immediately preceding the collision, with virtually no effective response during the preceding danger approach phase. This is highly detrimental to early risk screening. Although the DRAC indicator provides continuous numerical output throughout the process, its curve exhibits severe nonlinear oscillations, making it difficult to reliably quantify the evolution of risk trends. While the PODAR method in Fig. \ref{Curves of Various Metrics Under Safe and Critical Scenarios} (c) generates a continuous score, its peak characteristics at the collision threshold point are less pronounced than those of the method proposed in this paper, resulting in slightly insufficient discrimination capability. In contrast, the comprehensive risk potential energy depicted in Fig. \ref{Curves of Various Metrics Under Safe and Critical Scenarios} (d) demonstrates exceptional spatiotemporal continuity and foresight. In collision scenario B, our model acutely detects anomalies in the rear vehicle's speed and distance at t=0s, with the risk value initiating a monotonically increasing trend that accurately anticipates the accumulation of danger. As the vehicles approached, the risk potential energy climbed smoothly and rapidly, peaking near t=3s when the collision occurred, then rapidly declining as the vehicles separated. This characteristic of early detection, continuous evolution, and precise peak formation demonstrates that the model integrating the enhanced driver risk field with dynamic costs overcomes the threshold limitations of traditional kinematic indicators. It provides a supervised signal for hazardous scenario screening that not only aligns with physical laws but also possesses early warning capabilities.

\begin{figure*}[tp]
\centering
\includegraphics[width=0.95\textwidth]{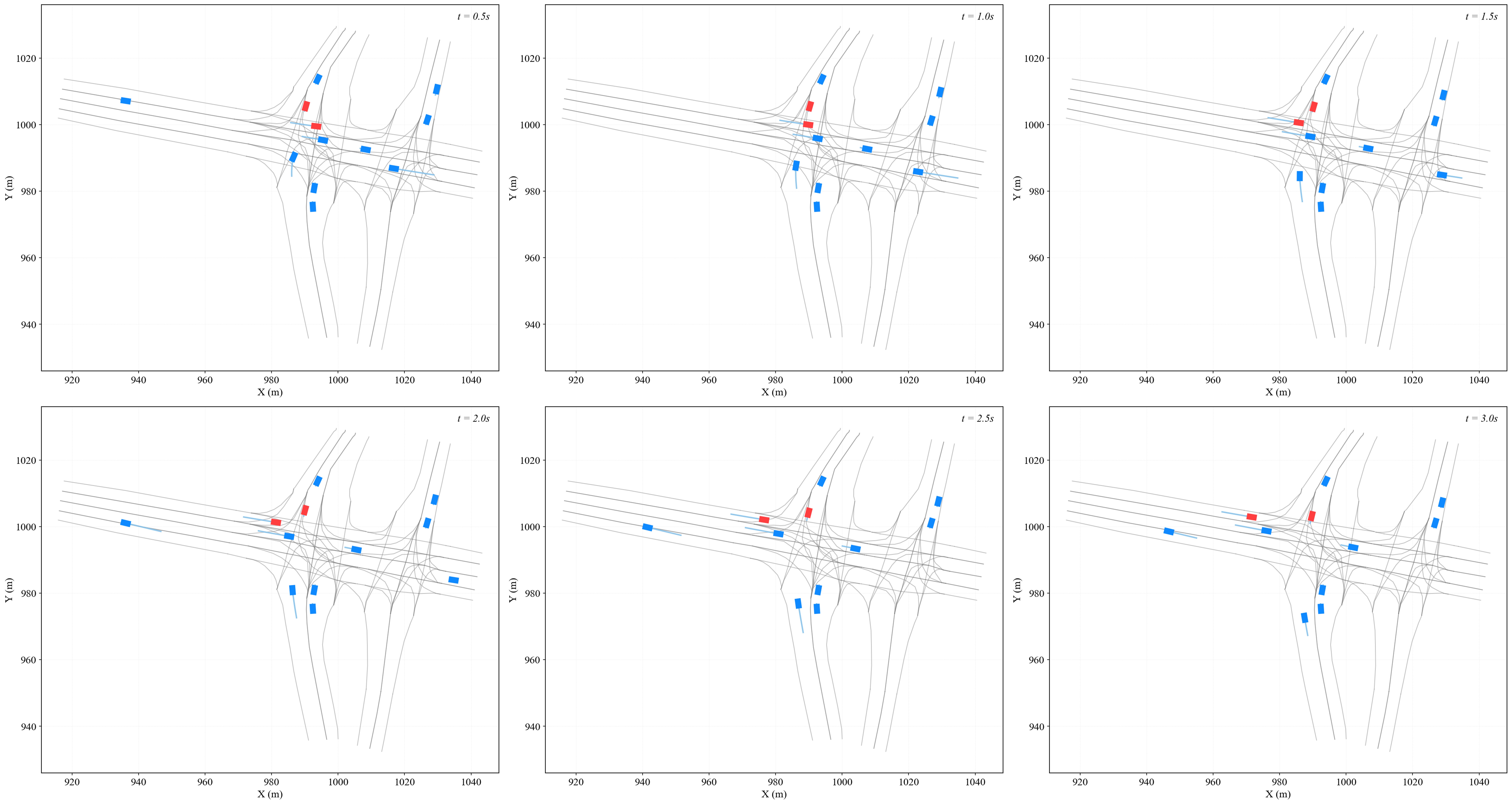}
\caption{Vehicle trajectories over 40 consecutive frames at an unsignalized intersection. A multi vehicle interaction example where vehicles approach and traverse the conflict zone, providing a visual reference for interpreting the corresponding risk evolution and peak risk moment.}
\label{Vehicle trajectories over 40 consecutive frames at an unsignalized intersection}
\end{figure*}

To further validate the model's effectiveness and accuracy in risk assessment across multiple interaction scenarios, we select specific interaction cases from the INTERACTION dataset to visually demonstrate the dynamic evolution of model risk over time. Fig. \ref{Vehicle trajectories over 40 consecutive frames at an unsignalized intersection} presents a schematic diagram of vehicle trajectories spanning 40 consecutive frames in an unsignalized intersection scenario.

As shown in Fig. \ref{Vehicle trajectories over 40 consecutive frames at an unsignalized intersection} During the initial phase, the red vehicle above gradually approaches the potential conflict zone at the intersection while two straight through vehicles are entering the intersection. As time progresses, the spatial distance between the straight through vehicles and the red vehicle rapidly decreases, with the degree of interaction continuously increasing. At t=1.5s, the red vehicle's risk perception zone significantly overlaps with the straight through vehicles, reaching the peak collision risk.

\begin{figure}[!t]
\centering
\includegraphics[width=0.48\textwidth]{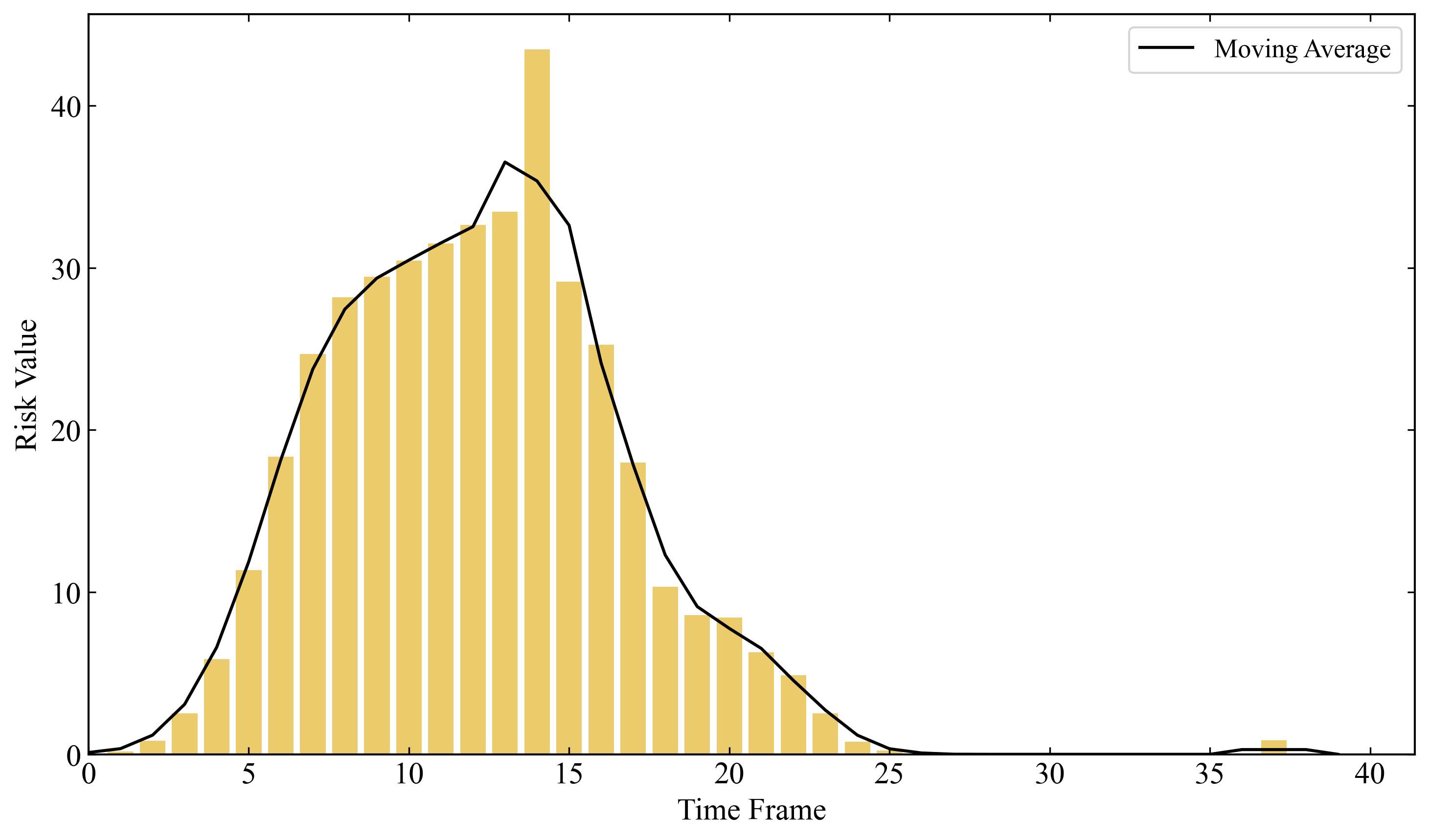}
\caption{Bar Chart of Risk Evolution for Red Vehicles at Intersections. Bar chart of risk over time for the red vehicle, showing a clear rise as the interaction intensifies and a rapid drop after vehicles leave the conflict zone, consistent with the trajectory level interaction process in Fig. 11.}
\label{Bar Chart of Risk Evolution for Red Vehicles at Intersections}
\end{figure}

The corresponding risk evolution curve is shown in Fig. \ref{Bar Chart of Risk Evolution for Red Vehicles at Intersections}. Near frame 15, the risk value for the red vehicle reaches its peak, closely matching the moment of maximum potential collision probability among the three vehicles in spatial position at t=1.5s in Fig. \ref{Vehicle trajectories over 40 consecutive frames at an unsignalized intersection}. Subsequently, as the straight-through vehicle exits the intersection conflict zone, the risk value rapidly drops back to a safe level.

This case demonstrates that the proposed risk modeling method accurately captures the rising and falling trends of risk during interactions. Its output temporal risk curve aligns with the actual interaction process. This indicates that the model not only successfully identifies high risk moments but also possesses the capability for continuous dynamic risk assessment, exhibiting strong predictive capability.

\subsection{Analysis of Scene Filtering Results}
To systematically evaluate the high value scene filtering capability of the risk modeling method proposed in this paper, we assess model performance using the area under the ROC curve (AUC), the area under the precision-recall curve (AP), and the Precision@K metric, respectively, for the risk labels in the FLUID dataset \cite{saito2015precision}, \cite{ceri2013introduction}. Specifically, AUC measures overall ranking capability, AP reflects identification performance under sparse hazardous event conditions, while Precision@K directly evaluates the model's practical effectiveness in hazardous segment screening tasks.

Additionally, to facilitate comparison with other proxy safety metrics, we select four representative indicators: PET, THW, DRAC, and PODAR \cite{chen2025podar}. To enable scale-equivalent comparison with the risk scores in this paper, the directionality of each metric is unified by taking the reciprocal of PET and THW, ensuring that higher values indicate greater risk.

\begin{figure}[!t]
\includegraphics[width=0.48\textwidth]{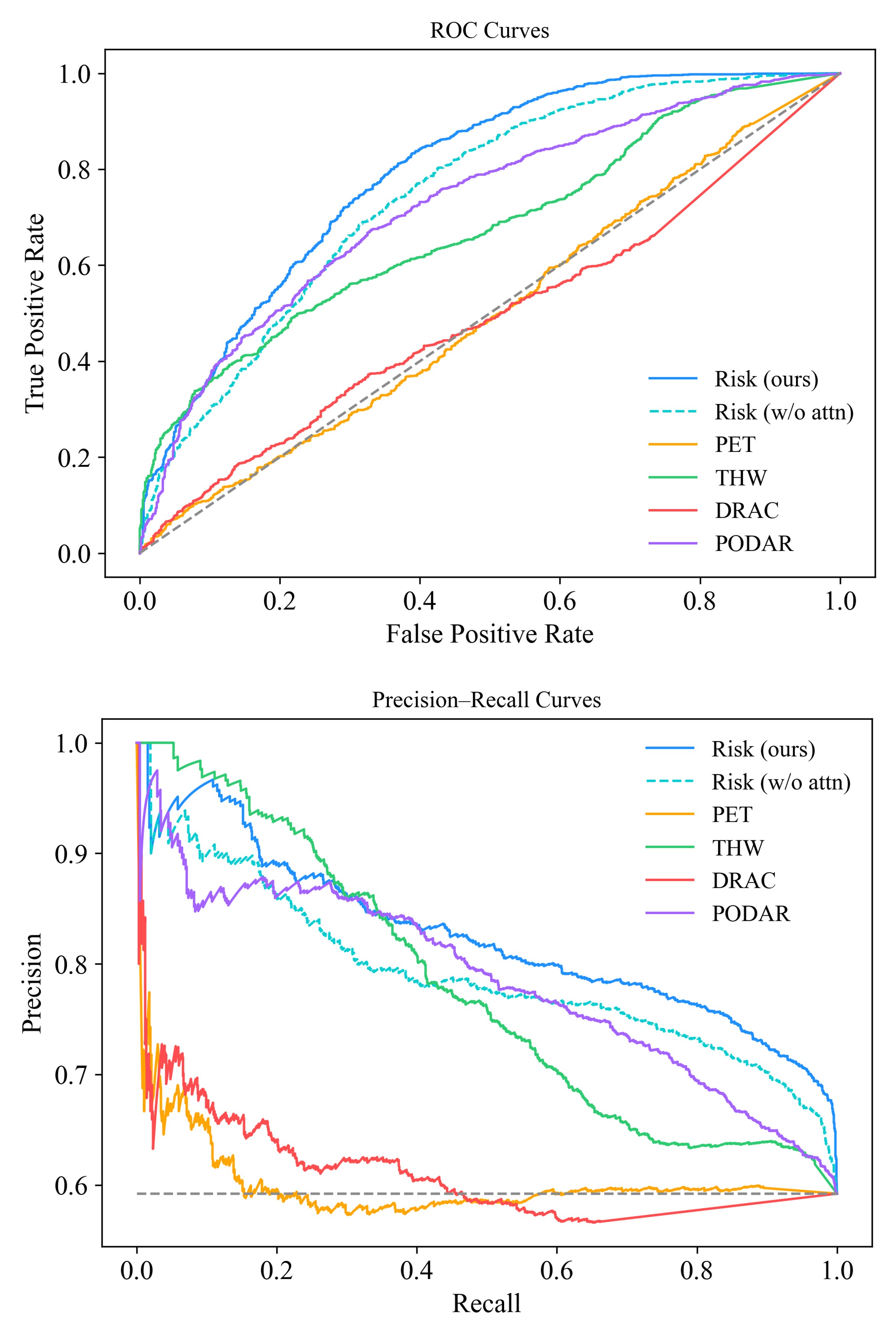}
\caption{ROC Curve and PR Curve for the FLUID Dataset. Screening performance comparison demonstrating that the proposed method achieves stronger global ranking (ROC) and better enrichment under long tail hazard sparsity (PR), validating its effectiveness for high risk priority screening.}
\label{ROC Curve and PR Curve for the FLUID Dataset}
\end{figure}

As shown in Fig. \ref{ROC Curve and PR Curve for the FLUID Dataset}, the model's ROC curve on the FLUID dataset significantly outperforms traditional security agent metrics. This demonstrates its ability to consistently rank hazardous samples ahead of safe ones, indicating robust global risk identification capabilities. The PR curve demonstrates that the model can still significantly enrich dangerous samples under the long-tail distribution of risk scenarios. In regions with lower recall rates, the accuracy remains at a high level, reflecting the model's application value in “high-risk priority” screening tasks.

\begin{table}[htbp]
  \centering
  \captionsetup{skip=0pt}
  \caption{Risk Identification Performance on the FLUID Dataset.}
  \begin{tabular}{
    w{l}{2.0cm}  
    w{c}{1.2cm}  
    w{c}{1.2cm}  
    w{c}{1.2cm}  
    w{c}{1.2cm}  
  }
    \toprule
    Method       & AUC       & AP        & P@100    & P@1000   \\
    \midrule
    PET          & 0.501     & 0.601     & 0.67     & 0.584    \\
    THW          & 0.676     & 0.777     & \textbf{0.98} & 0.732    \\
    DRAC         & 0.493     & 0.615     & 0.72     & 0.593    \\
    PODAR        & 0.726     & 0.785     & 0.89     & \underline{0.768} \\
    Ours(w/o attn) & \underline{0.749} & \underline{0.790} & 0.93     & \underline{0.768} \\
    Ours         & \textbf{0.792} & \textbf{0.825} & \underline{0.95} & \textbf{0.799} \\
    \bottomrule
  \end{tabular}
  \label{tab:method_compare}
\end{table}

As shown in Table \ref{tab:method_compare}, on the FLUID dataset, our model achieves an AUC of 0.792, indicating that when randomly sampling a pair of high risk and low risk samples, the model has a 79.2\% probability of correctly prioritizing the high risk sample over the low-risk one. Simultaneously, the average precision (AP) reached 0.825, significantly outperforming PET, THW, DRAC, and PODAR. This demonstrates the model's excellent hazardous sample enrichment capability and further proves its stable risk identification ability even under highly imbalanced sample distributions.

In addition, regarding Precision@K, the proposed model achieves 0.95 at P@100 and 0.799 at P@1000. This means that, when screening the top 1,000 most hazardous samples, 79.9\% of the selected events are truly dangerous. As a result, the manual screening workload is substantially reduced, demonstrating the method’s practical value for large scale hazardous scenario identification.

To verify the key role of the proposed risk–trajectory cross-attention module in mining potential hazardous scenarios, we conduct an ablation study. As shown in Table \ref{tab:method_compare}, Ours (w/o attn) denotes a baseline model that removes the cross-attention module in the decoder and retains only the multi-task parallel outputs. The results show that the full model with cross-attention achieves clear improvements across all key metrics. Specifically, AUC increases from 0.749 to 0.79 (about 4.3\%), and AP improves from 0.790 to 0.825. These gains strongly support the effectiveness of the proposed “implicit input–explicit output” mechanism. With an asymmetric cross-attention design, the trajectory branch is encouraged to extract features from the risk context that are highly related to interaction intent. This enhances sensitivity to high risk patterns and reduces the over smoothing tendency of purely data driven models on long tail samples.

\subsection{Runtime Efficiency}
To evaluate the real time performance of our model, we conduct experiments on a workstation equipped with an NVIDIA RTX 4090 GPU and an Intel Xeon Platinum 8375C CPU. The average inference time is approximately 38ms per frame, which meets the real time requirement of typical autonomous driving systems operating at 25 Hz. During inference, the peak GPU memory usage is about 4.6 GB. The computational demand is moderate, making the model compatible with most mainstream platforms, including the RTX 3060 and NVIDIA Jetson Orin.

\section{Conclusion}
We address the challenge of efficiently screening potentially hazardous scenarios from large scale data for autonomous driving testing. We propose a mechanism driven screening method based on risk quantification. Specifically, we construct a comprehensive risk function by integrating a Driver Risk Field with a dynamic cost model, which serves as the supervision signal. We further design a risk trajectory cross attention decoding scheme under an implicit input, explicit output paradigm. This design enables an autonomous pipeline that maps raw data inputs directly to scenario level risk scores for screening.

Experimental results show that, compared with traditional surrogate measures such as TTC, our risk modeling approach captures potential risk in a more continuous and fine grained manner, thereby providing high quality supervision for the screening task. The proposed risk aware prediction model also demonstrates strong capability in risk recognition and ranking on high conflict datasets such as FLUID. In addition, the workload recall analysis indicates that the method can substantially improve screening efficiency. The approach supports high speed inference and exhibits clear practical utility.

Future work will leverage the screened high risk scenarios as seed data. In combination with generative models, these seeds can be expanded into more diverse and higher risk edge cases, contributing to a more comprehensive test case library for autonomous driving.

\printbibliography

@article{sun2021scenario,
  title={Scenario-based test automation for highly automated vehicles: A review and paving the way for systematic safety assurance},
  author={Sun, Jian and Zhang, He and Zhou, Huajun and Yu, Rongjie and Tian, Ye},
  journal={IEEE transactions on intelligent transportation systems},
  volume={23},
  number={9},
  pages={14088--14103},
  year={2021},
  publisher={IEEE}
}

@article{bian2025search,
  title={Search-to-Crash: Generating safety-critical scenarios from in-depth crash data for testing autonomous vehicles},
  author={Bian, Jiang and Huang, Helai and Yu, Qianyuan and Zhou, Rui},
  journal={Energy},
  pages={137174},
  year={2025},
  publisher={Elsevier}
}

@inproceedings{watanabe2019scenario,
  title={Scenario mining for development of predictive safety functions},
  author={Watanabe, Hiroki and Tobisch, Lukas and Rost, Julia and Wallner, Johannes and Prokop, G{\"u}nther},
  booktitle={2019 IEEE International Conference on Vehicular Electronics and Safety (ICVES)},
  pages={1--7},
  year={2019},
  organization={IEEE}
}

@inproceedings{song2022scenario,
  title={A scenario distribution model for effective and efficient testing of autonomous driving systems},
  author={Song, Qunying and Runeson, Per and Persson, Stefan},
  booktitle={Proceedings of the 37th IEEE/ACM International Conference on Automated Software Engineering},
  pages={1--8},
  year={2022}
}

@article{tang2024scenario,
  title={Scenario-based accelerated testing for sotif in autonomous driving: A review},
  author={Tang, Lei and Wang, Ruijie and Liu, Zhanwen and Liang, Yunji and Niu, Yuanyuan and Zhu, Wei and Duan, Zongtao},
  journal={IEEE Internet of Things Journal},
  year={2024},
  publisher={IEEE}
}

@article{xu2025wod,
  title={Wod-e2e: Waymo open dataset for end-to-end driving in challenging long-tail scenarios},
  author={Xu, Runsheng and Lin, Hubert and Jeon, Wonseok and Feng, Hao and Zou, Yuliang and Sun, Liting and Gorman, John and Tolstaya, Ekaterina and Tang, Sarah and White, Brandyn and others},
  journal={arXiv preprint arXiv:2510.26125},
  year={2025}
}

@article{singh2024conflict,
  title={Conflict-Based safety evaluations at unsignalized intersections using surrogate safety measures},
  author={Singh, Dungar and Das, Pritikana and Ghosh, Indrajit},
  journal={Heliyon},
  volume={10},
  number={5},
  year={2024},
  publisher={Elsevier}
}

@article{kolekar2021risk,
  title={A risk field-based metric correlates with driver’s perceived risk in manual and automated driving: A test-track study},
  author={Kolekar, Sarvesh and Petermeijer, Bastiaan and Boer, Erwin and de Winter, Joost and Abbink, David},
  journal={Transportation research part C: emerging technologies},
  volume={133},
  pages={103428},
  year={2021},
  publisher={Elsevier}
}

@article{katariya2022deeptrack,
  title={Deeptrack: Lightweight deep learning for vehicle trajectory prediction in highways},
  author={Katariya, Vinit and Baharani, Mohammadreza and Morris, Nichole and Shoghli, Omidreza and Tabkhi, Hamed},
  journal={IEEE Transactions on Intelligent Transportation Systems},
  volume={23},
  number={10},
  pages={18927--18936},
  year={2022},
  publisher={IEEE}
}

@article{hui2022deep,
  title={Deep encoder--decoder-NN: A deep learning-based autonomous vehicle trajectory prediction and correction model},
  author={Hui, Fei and Wei, Cheng and ShangGuan, Wei and Ando, Ryosuke and Fang, Shan},
  journal={Physica A: Statistical Mechanics and its Applications},
  volume={593},
  pages={126869},
  year={2022},
  publisher={Elsevier}
}

@inproceedings{xie2024advdiffuser,
  title={Advdiffuser: Generating adversarial safety-critical driving scenarios via guided diffusion},
  author={Xie, Yuting and Guo, Xianda and Wang, Cong and Liu, Kunhua and Chen, Long},
  booktitle={2024 IEEE/RSJ International Conference on Intelligent Robots and Systems (IROS)},
  pages={9983--9989},
  year={2024},
  organization={IEEE}
}

@article{tafidis2023application,
  title={Application of surrogate safety measures in higher levels of automated vehicles simulation studies: A review of the state of the practice},
  author={Tafidis, Pavlos and Pirdavani, Ali},
  journal={Traffic injury prevention},
  volume={24},
  number={3},
  pages={279--286},
  year={2023},
  publisher={Taylor \& Francis}
}

@article{wang2022acceleration,
  title={Acceleration-based collision criticality metric for holistic online safety assessment in automated driving},
  author={Wang, Cheng and Popp, Christoph and Winner, Hermann},
  journal={IEEE Access},
  volume={10},
  pages={70662--70674},
  year={2022},
  publisher={IEEE}
}

@article{chang2018integration,
  title={Integration of speed and time for estimating time to contact},
  author={Chang, Chia-Jung and Jazayeri, Mehrdad},
  journal={Proceedings of the National Academy of Sciences},
  volume={115},
  number={12},
  pages={E2879--E2887},
  year={2018},
  publisher={National Academy of Sciences}
}

@article{cheng2025emergency,
  title={Emergency Index (EI): A two-dimensional surrogate safety measure considering vehicles’ interaction depth},
  author={Cheng, Hao and Jiang, Yanbo and Zhang, Hailun and Chen, Keyu and Huang, Heye and Xu, Shaobing and Wang, Jianqiang and Zheng, Sifa},
  journal={Transportation Research Part C: Emerging Technologies},
  volume={171},
  pages={104981},
  year={2025},
  publisher={Elsevier}
}

@inproceedings{lyu2024risk,
  title={Risk-based Socially-Compliant Behavior Planning for Autonomous Driving},
  author={Lyu, Yiwei and Luo, Wenhao and Dolan, John M},
  booktitle={2024 American Control Conference (ACC)},
  pages={3827--3832},
  year={2024},
  organization={IEEE}
}

@article{ma2023real,
  title={Real-time risk assessment model for multi-vehicle interaction of connected and autonomous vehicles in weaving area based on risk potential field},
  author={Ma, Yanli and Dong, Fangqi and Yin, Biqing and Lou, Yining},
  journal={Physica A: Statistical Mechanics and its Applications},
  volume={620},
  pages={128725},
  year={2023},
  publisher={Elsevier}
}

@article{wang2016driving,
  title={Driving safety field theory modeling and its application in pre-collision warning system},
  author={Wang, Jianqiang and Wu, Jian and Zheng, Xunjia and Ni, Daiheng and Li, Keqiang},
  journal={Transportation research part C: emerging technologies},
  volume={72},
  pages={306--324},
  year={2016},
  publisher={Elsevier}
}

@article{huang2020probabilistic,
  title={A probabilistic risk assessment framework considering lane-changing behavior interaction},
  author={Huang, Heye and Wang, Jianqiang and Fei, Cong and Zheng, Xunjia and Yang, Yibin and Liu, Jinxin and Wu, Xiangbin and Xu, Qing},
  journal={Science China Information Sciences},
  volume={63},
  number={9},
  pages={190203},
  year={2020},
  publisher={Springer}
}

@article{xu2023driving,
  title={Driving risk field and control strategies for autonomous vehicles at a signalized intersection},
  author={Xu, Hui and Wu, Jianping},
  journal={Journal of advanced transportation},
  volume={2023},
  number={1},
  pages={8072495},
  year={2023},
  publisher={Wiley Online Library}
}

@inproceedings{schubert2008comparison,
  title={Comparison and evaluation of advanced motion models for vehicle tracking},
  author={Schubert, Robin and Richter, Eric and Wanielik, Gerd},
  booktitle={2008 11th international conference on information fusion},
  pages={1--6},
  year={2008},
  organization={IEEE}
}

@inproceedings{barth2008will,
  title={Where will the oncoming vehicle be the next second?},
  author={Barth, Alexander and Franke, Uwe},
  booktitle={2008 IEEE Intelligent Vehicles Symposium},
  pages={1068--1073},
  year={2008},
  organization={IEEE}
}

@inproceedings{lytrivis2008cooperative,
  title={Cooperative path prediction in vehicular environments},
  author={Lytrivis, Panagiotis and Thomaidis, George and Amditis, Angelos},
  booktitle={2008 11th international IEEE conference on intelligent transportation systems},
  pages={803--808},
  year={2008},
  organization={IEEE}
}

@article{zhang2017method,
  title={A method for connected vehicle trajectory prediction and collision warning algorithm based on V2V communication},
  author={Zhang, Ruifeng and Cao, Libo and Bao, Shan and Tan, Jianjie},
  journal={International Journal of Crashworthiness},
  volume={22},
  number={1},
  pages={15--25},
  year={2017},
  publisher={Taylor \& Francis}
}

@inproceedings{schulz2019learning,
  title={Learning interaction-aware probabilistic driver behavior models from urban scenarios},
  author={Schulz, Jens and Hubmann, Constantin and Morin, Nikolai and L{\"o}chner, Julian and Burschka, Darius},
  booktitle={2019 IEEE intelligent vehicles symposium (iv)},
  pages={1326--1333},
  year={2019},
  organization={IEEE}
}

@inproceedings{li2022autonomous,
  title={Autonomous driving behavior prediction method based on improved hidden Markov model},
  author={Li, Te and Chen, Liqiong and Wang, Ying},
  booktitle={2022 IEEE 25th International Conference on Computer Supported Cooperative Work in Design (CSCWD)},
  pages={758--762},
  year={2022},
  organization={IEEE}
}

@article{qiao2014self,
  title={A self-adaptive parameter selection trajectory prediction approach via hidden Markov models},
  author={Qiao, Shaojie and Shen, Dayong and Wang, Xiaoteng and Han, Nan and Zhu, William},
  journal={IEEE Transactions on Intelligent Transportation Systems},
  volume={16},
  number={1},
  pages={284--296},
  year={2014},
  publisher={IEEE}
}

@article{dai2019modeling,
  title={Modeling vehicle interactions via modified LSTM models for trajectory prediction},
  author={Dai, Shengzhe and Li, Li and Li, Zhiheng},
  journal={Ieee Access},
  volume={7},
  pages={38287--38296},
  year={2019},
  publisher={IEEE}
}

@inproceedings{kim2017probabilistic,
  title={Probabilistic vehicle trajectory prediction over occupancy grid map via recurrent neural network},
  author={Kim, ByeoungDo and Kang, Chang Mook and Kim, Jaekyum and Lee, Seung Hi and Chung, Chung Choo and Choi, Jun Won},
  booktitle={2017 IEEE 20Th international conference on intelligent transportation systems (ITSC)},
  pages={399--404},
  year={2017},
  organization={IEEE}
}

@inproceedings{xu2022adaptive,
  title={Adaptive trajectory prediction via transferable gnn},
  author={Xu, Yi and Wang, Lichen and Wang, Yizhou and Fu, Yun},
  booktitle={Proceedings of the IEEE/CVF conference on computer vision and pattern recognition},
  pages={6520--6531},
  year={2022}
}

@inproceedings{gao2020vectornet,
  title={Vectornet: Encoding hd maps and agent dynamics from vectorized representation},
  author={Gao, Jiyang and Sun, Chen and Zhao, Hang and Shen, Yi and Anguelov, Dragomir and Li, Congcong and Schmid, Cordelia},
  booktitle={Proceedings of the IEEE/CVF conference on computer vision and pattern recognition},
  pages={11525--11533},
  year={2020}
}

@inproceedings{zhu2022spatio,
  title={A spatio-temporal graph transformer network for multi-pedestrain trajectory prediction},
  author={Zhu, Jingfei and Lian, Zhichao and Jiang, Zhukai},
  booktitle={2022 IEEE Intl Conf on Dependable, Autonomic and Secure Computing, Intl Conf on Pervasive Intelligence and Computing, Intl Conf on Cloud and Big Data Computing, Intl Conf on Cyber Science and Technology Congress (DASC/PiCom/CBDCom/CyberSciTech)},
  pages={1--5},
  year={2022},
  organization={IEEE}
}

@inproceedings{wang2023prophnet,
  title={Prophnet: Efficient agent-centric motion forecasting with anchor-informed proposals},
  author={Wang, Xishun and Su, Tong and Da, Fang and Yang, Xiaodong},
  booktitle={Proceedings of the IEEE/CVF conference on computer vision and pattern recognition},
  pages={21995--22003},
  year={2023}
}

@inproceedings{tang2024hpnet,
  title={Hpnet: Dynamic trajectory forecasting with historical prediction attention},
  author={Tang, Xiaolong and Kan, Meina and Shan, Shiguang and Ji, Zhilong and Bai, Jinfeng and Chen, Xilin},
  booktitle={Proceedings of the IEEE/CVF conference on computer vision and pattern recognition},
  pages={15261--15270},
  year={2024}
}

@article{lee2016stochastic,
  title={Stochastic multiple choice learning for training diverse deep ensembles},
  author={Lee, Stefan and Purushwalkam Shiva Prakash, Senthil and Cogswell, Michael and Ranjan, Viresh and Crandall, David and Batra, Dhruv},
  journal={Advances in Neural Information Processing Systems},
  volume={29},
  year={2016}
}

@article{gettman2003surrogate,
  title={Surrogate safety measures from traffic simulation models},
  author={Gettman, Douglas and Head, Larry},
  journal={Transportation research record},
  volume={1840},
  number={1},
  pages={104--115},
  year={2003},
  publisher={SAGE Publications Sage CA: Los Angeles, CA}
}

@article{kolekar2020human,
  title={Human-like driving behaviour emerges from a risk-based driver model},
  author={Kolekar, Sarvesh and De Winter, Joost and Abbink, David},
  journal={Nature communications},
  volume={11},
  number={1},
  pages={4850},
  year={2020},
  publisher={Nature Publishing Group UK London}
}

@article{zhan2019interaction,
  title={Interaction dataset: An international, adversarial and cooperative motion dataset in interactive driving scenarios with semantic maps},
  author={Zhan, Wei and Sun, Liting and Wang, Di and Shi, Haojie and Clausse, Aubrey and Naumann, Maximilian and Kummerle, Julius and Konigshof, Hendrik and Stiller, Christoph and de La Fortelle, Arnaud and others},
  journal={arXiv preprint arXiv:1910.03088},
  year={2019}
}

@article{chen2025fluid,
  title={FLUID: A Fine-Grained Lightweight Urban Signalized-Intersection Dataset of Dense Conflict Trajectories},
  author={Chen, Yiyang and Wu, Zhigang and Zheng, Guohong and Wu, Xuesong and Xu, Liwen and Tang, Haoyuan and He, Zhaocheng and Zeng, Haipeng},
  journal={arXiv preprint arXiv:2509.00497},
  year={2025}
}

@article{saito2015precision,
  title={The precision-recall plot is more informative than the ROC plot when evaluating binary classifiers on imbalanced datasets},
  author={Saito, Takaya and Rehmsmeier, Marc},
  journal={PloS one},
  volume={10},
  number={3},
  pages={e0118432},
  year={2015},
  publisher={Public Library of Science San Francisco, CA USA}
}

@incollection{ceri2013introduction,
  title={An introduction to information retrieval},
  author={Ceri, Stefano and Bozzon, Alessandro and Brambilla, Marco and Della Valle, Emanuele and Fraternali, Piero and Quarteroni, Silvia},
  booktitle={Web information retrieval},
  pages={3--11},
  year={2013},
  publisher={Springer}
}

@article{chen2025podar,
  title={PODAR: A Collision Risk Model Offering Valid Signals for Vehicular Interactions},
  author={Chen, Chen and Liu, Zhipeng and Liang, Weiqiang and Wang, Mingming and Lan, Zhiqian and Zhan, Guojian and Lyu, Yao and Nie, Bingbing and Li, Shengbo Eben},
  journal={IEEE Intelligent Transportation Systems Magazine},
  year={2025},
  publisher={IEEE}
}

\vspace{1em}

\newpage
\vfill
\end{document}